%% file: revision_v2.tex
\documentclass[final,3p,times,twocolumn]{elsarticle}
\usepackage{amsfonts}
\usepackage{algorithmic}
\usepackage{algorithm}
\usepackage{array}
\usepackage[table]{xcolor}
\usepackage{textcomp}
\usepackage{stfloats}
\usepackage{url}
\usepackage{verbatim}
\usepackage{graphicx}
\usepackage{cite}
\usepackage{booktabs} 
\usepackage{multirow} 
\usepackage{amssymb}
\usepackage{amsmath}
\usepackage{dsfont}
\usepackage{pifont}
\usepackage{bm}
\usepackage{enumitem}
\usepackage{subcaption}
\usepackage[export]{adjustbox}
\newcommand{\revise}[1]{\textcolor{black}{#1}}
\newcommand{\srevise}[1]{\textcolor{black}{#1}}

\newcommand{\shortname}{SfMamba}
\newcommand{\modela}{Channel-wise Visual State-Space}
\newcommand{\shorta}{Ch-VSS}
\newcommand{\modelb}{Semantic-Consistent Shuffle}
\newcommand{\shortb}{SCS}
\definecolor{ForestGreen}{rgb}{0.13, 0.55, 0.13}

\newcommand{\stdvu}[1]{{\footnotesize\color{red}$\uparrow$~\color{darkgray}{#1}}}

\newcommand{\stdvddown}[1]{{\footnotesize \color{ForestGreen}$\downarrow$~\color{darkgray}{#1}}}
\newcommand{\stdvflat}[1]{{\footnotesize\color{gray}$-$~\color{darkgray}{#1}}}

\newcommand{\ie}{\text{i}.\text{e}.}
\newcommand{\eg}{\text{e}.\text{g}.}

\begin{document}
\begin{frontmatter}
\title{SfMamba: Efficient Source-Free Domain Adaptation via Selective Scan Modeling}

\author[1]{Xi Chen}
\affiliation[1]{organization={School of Computer Science and Technology, Harbin Institute of Technology}, city={Harbin}, postcode={150001}, country={China}}

\author[1]{Hongxun Yao\corref{cor1}}
\cortext[cor1]{Corresponding Author}
\author[2]{Sicheng Zhao}
\affiliation[2]{organization={Department of Psychological and Cognitive Sciences, Tsinghua University, Beijing},  postcode={100084}, country={China}}

\author[1]{Jiankun Zhu}
\author[1]{Jing Jiang}
\author[1]{Kui Jiang}

\nonumnote{
E-mail address: xichen98cn@gmail.com (X. Chen), h.yao@hit.edu.cn (H. Yao).
}

\begin{abstract}
Source-free domain adaptation (SFDA) tackles the critical challenge of adapting source-pretrained models to unlabeled target domains without access to source data, overcoming data privacy and storage limitations in real-world applications. However, existing SFDA approaches struggle with the trade-off between perception field and computational efficiency in domain-invariant feature learning. Recently, Mamba has offered a promising solution through its selective scan mechanism, which enables long-range dependency modeling with linear complexity. However, the Visual Mamba (\ie, VMamba) remains limited in capturing channel-wise frequency characteristics critical for domain alignment and maintaining spatial robustness under significant domain shifts. To address these, we propose a framework called \shortname{} to fully explore the stable dependency in source-free model transfer. \shortname{} introduces \modela{} block that enables channel-sequence scanning for domain-invariant feature extraction. In addition, \shortname{} involves a \modelb{} strategy that disrupts background patch sequences in 2D selective scan while preserving prediction consistency to mitigate error accumulation. Comprehensive evaluations across multiple benchmarks show that \shortname{} achieves consistently stronger performance than existing methods while maintaining favorable parameter efficiency, offering a practical solution for SFDA. Our code is available at \textit{https://github.com/chenxi52/SfMamba}. 
\end{abstract}

\end{frontmatter}

\section{Introduction}
Domain adaptation (DA)~\citep{DANN,CDTrans} aims to mitigate the performance degradation of machine learning models when deployed in a target domain distinct from the source domain on which they were trained. Traditional DA methods typically rely on concurrent access to both labeled source data and unlabeled target data to align their distributions. However, this assumption is often impractical in real-world scenarios due to privacy concerns, computational constraints, or legal restrictions that prevent source data sharing. Source-free domain adaptation (SFDA)~\citep{SHOT, A2Net,NRC} addresses this challenge by adapting a trained source model to the target domain without revisiting the original source data during adaptation. This paradigm shifts the focus to leveraging the model’s intrinsic knowledge (\eg, feature representations or batch statistics) to transfer the model to the target domain. 

\begin{figure}[!t]
    \centering
\subfloat[]{\includegraphics[width=.98\linewidth]{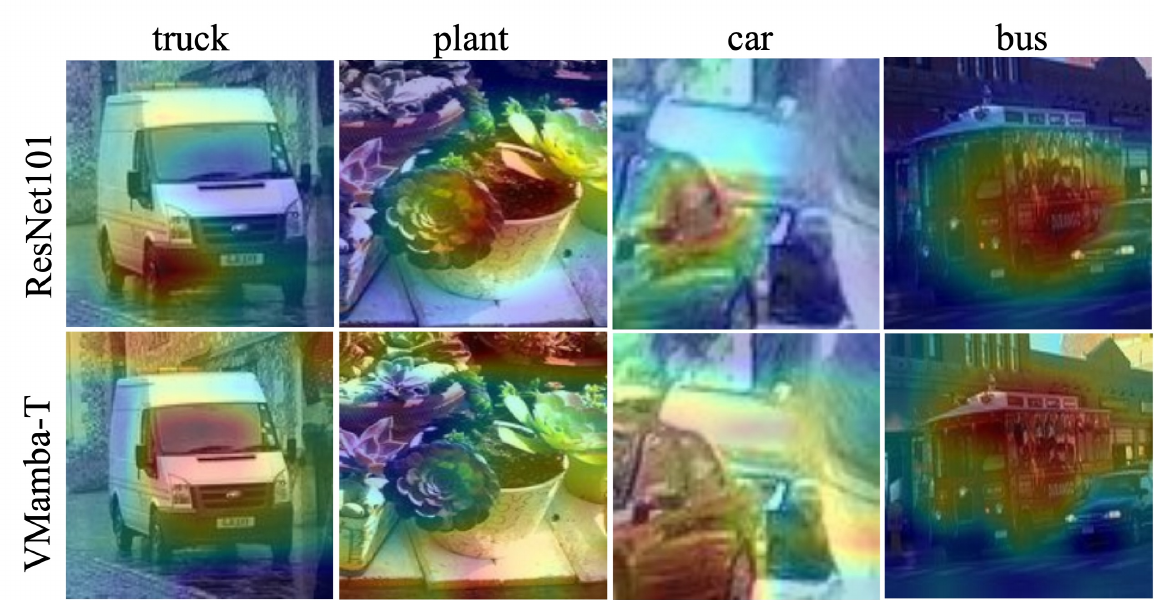}
\label{fig:cam_vsRes}}
\vspace{1pt}
\subfloat[]{
\includegraphics[width=.98\linewidth]{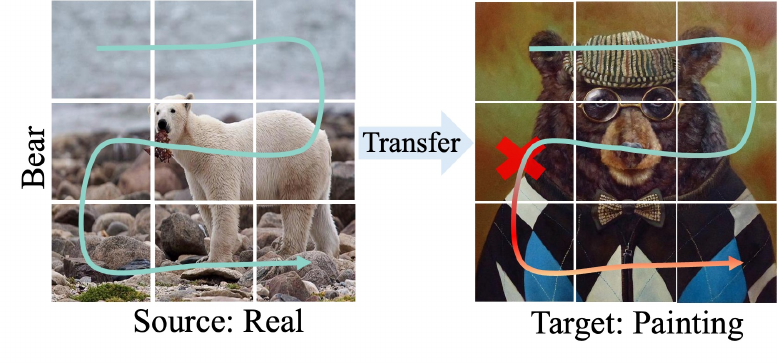}
\label{fig:spatial_scanning}
}
    \caption{(a) Grad-CAM~\citep{Grad-cam} visualizations from SHOT using ResNet-101 versus VMamba-Tiny on VisDA-C~\citep{Visda-c}. While ResNet exhibits a constrained receptive field with central bias, VMamba achieves broader spatial awareness. 
    (b) Conceptual illustration of how domain shift induces error propagation in state-space models. In the Real $\to$ Painting task (DomainNet-126~\citep{domainnet126}), spurious sequence correlations from the source domain misguide state evolution, causing errors to accumulate in the target domain.
}
\end{figure}

Previous SFDA works have primarily relied on convolutional neural networks (CNN)~\citep{Resnet} or Vision Transformers (ViT)~\citep{vit} as feature backbone. 
CNNs are inherently limited by their local receptive fields, which can hinder effective domain alignment. As illustrated in Fig.~\ref{fig:cam_vsRes}, GradCAM visualizations from a ResNet-based SFDA model show a focus on central image regions, failing to capture broader contextual information.
Although ViTs~\citep{Dsit,TransDA} mitigate this through global attention mechanisms, they incur quadratic computational complexity, which is prohibitive for many applications.  
In contrast, the selective state-space model (SSM) Mamba~\citep{mamba} formulates a dynamic system with a parallel selective scan operation, enabling long-range dependency modeling with linear-time complexity. These characteristics make it practical for real-world SFDA applications, where capturing extensive domain-invariant dependencies while maintaining computational efficiency is crucial. 
Quantitative analysis substantiates these advantages: Fig.~\ref{fig:cam_vsRes} demonstrates that Visual Mamba~\citep{vmamba} generates activation maps that more comprehensively highlight complete object structures compared to ResNet.

\srevise{Despite these promising characteristics, we observe that the standard Visual Mamba, when applied to SFDA, exhibits suboptimal attention and fails to robustly capture domain-invariant semantic object structures (Fig.~\ref{fig:cam_vsRes}). This highlights a need for enhanced feature conditioning within the Mamba paradigm for domain adaptation.}
The second limitation arises from the recurrent nature of the state-space model itself. As illustrated in Fig.~\ref{fig:spatial_scanning}, the sequence modeling introduces an inherent vulnerability in source-pretraining: entangle semantic content with incidental spatial co-occurrences in the source domain's scanning sequence. The model learns to rely not on the intrinsic features of the primary object, but on a specific configuration of spatial context that is statistically prevalent in the source domain. For instance, a model may learn to identify a "bear" not by its distinctive physical features, but by the consistent co-occurrence of environmental textures in its vicinity, such as rocky terrain or dense foliage. This learned association forms a robust shortcut for the source domain. However, during transfer to a target domain -- the artistic \textit{Painting}, these incidental spatial co-occurrences often break down. 
Without a built-in damping mechanism, estimation errors and noisy signals from the target domain can perpetuate and amplify through the hidden states during spatial scanning, ultimately degrading adaptation performance in the final state.

To this end, we propose \shortname{} to advance SFDA tasks with the Mamba framework. \srevise{Drawing inspiration from the effective channel-interaction principles in CNN-based domain adaptation~\citep{GDCAN,DCAN,G-SFDA}, which allows the model to prioritize semantically meaningful, frequency-stable features~\citep{FDA,FcaNet} crucial for cross-domain generalization, we conceptualize and implement a novel adaptation of this principle for Mamba's sequential modeling framework. Specifically, we incorporate \modela{} (\shorta) blocks following the feature backbone, where we utilize the original selective scanning mechanism in Mamba and employ bidirectional channel scanning on features along their channel dimension. This design enables dynamic, input-dependent modeling of channel correlations, directly facilitating the learning of domain-invariant frequency characteristics. }
Furthermore, to combat error accumulation, we propose a \modelb{} (\shortb) strategy, which employs a patch-perturbation during visual mamba's 2D selective scanning, selectively shuffling background patches while maintaining prediction invariance through consistency regularization.
By simultaneously enhancing learning in the channel-wise and spatial dimensions, \shortname{} achieves a superior parameter-accuracy trade-off in SFDA.

Our contributions are threefold:
\begin{itemize}
    \item We propose \shortname{}, the first efficient Mamba-based SFDA framework that leverages state-space models' long-range modeling capabilities with linear complexity.
    \item We design the \modela{} block for channel-wise domain-invariant feature learning and the \modelb{} strategy for robust spatial sequence processing, effectively tailoring the Mamba architecture to the demands of source-free adaptation.
    \item Comprehensive experiments on four benchmarks demonstrate that \shortname{} consistently outperforms existing methods.
\end{itemize}

The remainder of this paper is organized as follows. Section~\ref{sec:rela_work} reviews related work on traditional unsupervised domain adaptation, source-free domain adaptation, and the application of Mamba in vision tasks. Section~\ref{sec:method} formulates the SFDA problem and presents the necessary preliminaries along with our proposed approach. Section~\ref{sec:exp} details the experimental setup, reports the main results, and provides comprehensive ablation studies. \srevise{Section~\ref{sec:limit} discusses the limitations of our work and outlines potential directions for future research.} Finally, Section~\ref{sec:conclusion} concludes the paper.

\section{Related Work}
\label{sec:rela_work}
\subsection{Unsupervised Domain Adaptation}
Unsupervised Domain Adaptation (UDA) aims to mitigate the distribution shift between source and target domains. Existing approaches can be broadly divided into two main categories.
The first category focuses on learning domain-invariant representations by explicitly minimizing distribution divergence through feature transformation. For instance, some methods align domains by minimizing statistical measures such as Maximum Mean Discrepancy~\citep{MMD}, Margin Disparity Discrepancy~\citep{MDD}, or Wasserstein Divergence~\citep{RHWD}. Others, like CORA~\citep{CORAL} and its variants, perform correlation alignment of feature distributions. Adversarial methods, including DANN~\citep{DANN} and its successors RADA-prompt~\citep{RADA-prompt}, learn transferable features by introducing a domain discriminator and applying gradient reversal to confuse domain-specific characteristics.
The second category adopts self-learning strategies, where pseudo-labels are generated for unlabeled target data based on source-trained models, followed by iterative self-training or clustering-based refinement. Representative works include SRDC~\citep{SRDC}, which leverages discriminative clustering to reveal intrinsic target structures, and RSDA~\citep{RSDA}, which introduces robust pseudo-labeling in a spherical feature space. SimNet~\citep{SimNet}, on the other hand, performs classification by measuring similarity to category-wise prototypes.
Recent advances have further expanded UDA techniques. For example, pre-trained models have been widely adopted for better transferability~\citep{DAPL}, and generative approaches such as diffusion models are increasingly explored for cross-domain data augmentation or feature alignment~\citep{DAD}.
It is important to note, however, that UDA methods assume full access to labeled source data during training—an assumption that does not hold in the SFDA setting considered in this work.

\subsection{Source-Free Domain Adaptation} 
Source-free domain adaptation (SFDA) addresses the challenging scenario where a pre-trained source model must adapt to an unlabeled target domain without access to source data. \revise{This paradigm of learning under data constraints aligns with broader ``source-absent'' or self-supervised learning practices in vision. For example, hyperspectral image super-resolution methods~\citep{Model-informed, Enhanced_deep} demonstrate that images can be recovered without paired high-resolution supervision by leveraging model-informed or physical priors. Similarly, classical image restoration approaches such as Noise2Noise~\citep{noise2noise} show that reliable denoising is possible using only independently corrupted observations. These examples illustrate that robust statistical structure can often be extracted without explicit source–target data correspondence.} 

Existing SFDA approaches mainly employ two types of backbone architectures.
CNN-based methods primarily operate by exploring the feature representation space. SHOT~\citep{SHOT} establishes a foundational approach by generating pseudo-labels through feature clustering. Building on this, UPA~\citep{UPA} develops a more robust framework by analyzing both neighbor distribution and label consistency for noise filtering. NRC~\citep{NRC} enhances class-consistency through neighborhood affinity in the latent space, while NAMI~\citep{NAMI} further advances this direction by maximizing mutual information between target samples and their feature-space neighbors.
ViT-based methods leverage the transformer's cross-attention capabilities. DSiT~\citep{Dsit} innovates by extracting domain-specific and task-specific factors through query learning in the self-attention mechanism. TransDA~\citep{TransDA} incorporates transformer blocks after the feature extractor to exploit long-range attention for improved target generalization. C-SFTrans~\citep{C-SFTrans} introduces an attention-head selection mechanism to disentangle task-related features from contextual factors in the multi-head attention structure. \revise{DPC finds the attention maps region of interest is related to prediction, therefore adjusts the distribution of the attention matrix to involve more transferable patterns and introduces attention-induced sample mixture to expand target sample diversity. In contrast to these approaches, our work leverages a state space model for efficient long-range dependency modeling to capture domain-invariant transferable features for model transfer.}

\subsection{Mamba in Vision Task}
State Space Models (SSMs) have shown significant effectiveness in capturing dynamic dependencies within sequential data, particularly in language modeling.
Mamba~\citep{mamba} further advances SSMs by introducing data-dependent SSM layers and a selection mechanism through parallel scan. Recent vision adaptations such as Vim~\citep{vim} and VMamba~\citep{vmamba} leverage Mamba's linear computational complexity to process high-dimensional visual data, employing sophisticated scanning strategies to capture spatial relationships across image patches.
The Mamba architecture has been extended to diverse vision applications. Freqmamba~\citep{freqmamba} addresses image deraining through frequency-domain correlation modeling, while Style Mamba-Transformer~\citep{StyleMamba} enhances content preservation in text-style transfer tasks. MDAF~\citep{MDAF} addresses long-range dependency issues in image super-resolution, and in domain generalization, DGMamba~\citep{dgmamba} improves cross-domain robustness through domain-specific state suppression. DAMamba~\citep{DAMamba} focuses on domain adaptation in object detection by modeling domain variations, and \revise{DATMamba~\citep{DATMamba} employs Mamba-MLP-Attention-MLP architectural stage after the ResNet feature backbone for global context modeling. Unlike these approaches, our work provides the first systematic examination of a pure Mamba backbone's limitations under SFDA constraints. We specifically investigate how its sequential scanning mechanism affects domain-invariant learning when source data is inaccessible.}

\section{Method}
\label{sec:method}
\begin{figure*}[!th]
    \centering
\includegraphics[width=0.85\textwidth]{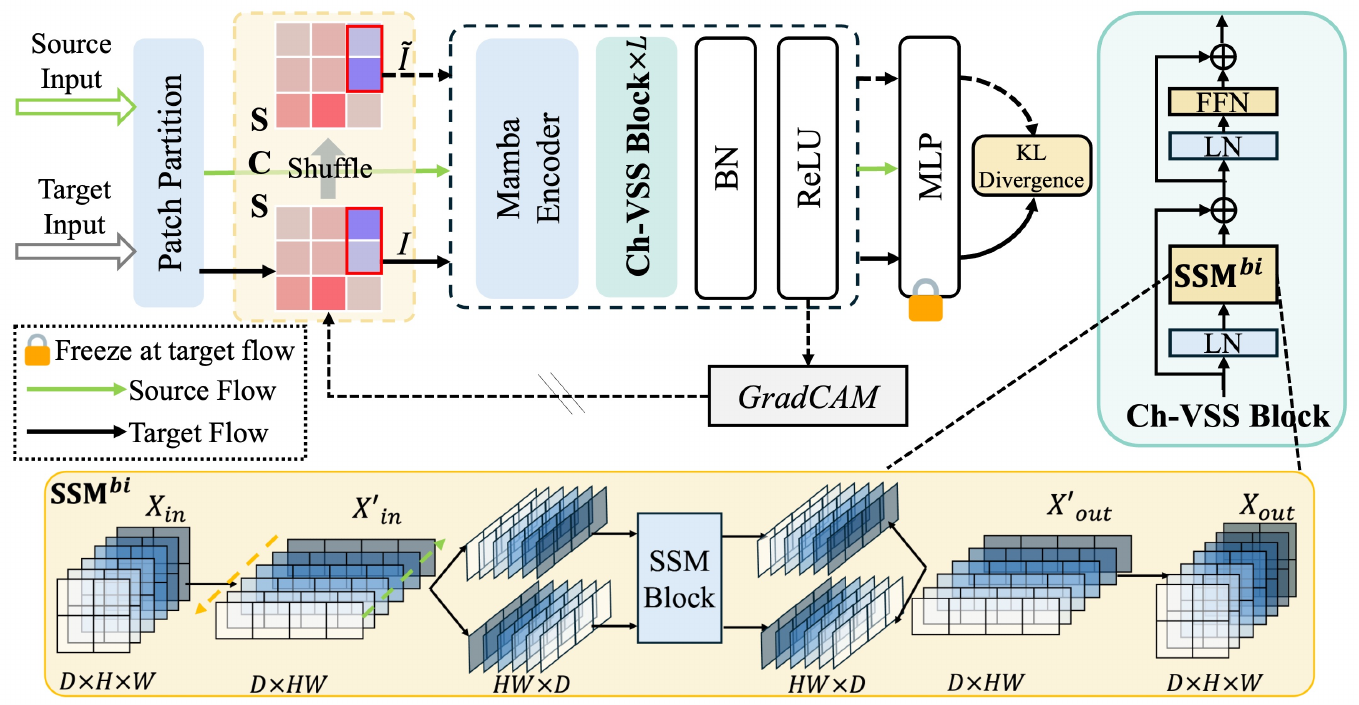}
    \caption{Overview of \shortname{} (left). The backbone adopts the encoder from VMamba~\citep{vmamba}, followed by an $L\times$ \modela{} (\shorta) block (detailed on the right) to perform bidirectional channel scanning. During target adaptation, the input $I$ is processed to generate a Grad-CAM activation map~\citep{Grad-cam}, where low-activation patches are identified as background and shuffled ($\tilde I$) via the \modelb{} (\shortb) strategy. Predictions from both original and perturbed sequences are regularized through KL divergence minimization. }
    \label{fig:frame}
\end{figure*}

\subsection{Problem Setting}
In SFDA, the objective is to train a model on labeled source data $\mathcal D_s={(x_s^i,y_s^i)}_{i=1}^{n_s}\in (\mathcal X_s, \mathcal Y)$ such that it can effectively transfer to an unlabeled target domain $\mathcal D_t=\{x_t^i\}_{i=1}^{n_t} \in \mathcal X_t$ sharing the same label space $\mathcal{Y}$, while maintaining the constraint of no source data access during adaptation. Let $C=|\mathcal Y|$ be the number of classes. The model $\theta=f\cdot g$, where $g: \mathcal X\xrightarrow{}\mathbb R^D$ denotes the feature extractor and $f:\mathbb R^D\xrightarrow{} \mathcal Y$ represents a multiple-layer perception (MLP) classifier. 
During the initial source training phase, $\theta_s$ is trained to minimize the empirical risk over $\mathcal D_s$, establishing initial feature representations and decision boundaries. During target adaptation, model $\theta_t$ is initialized with $\theta_s$. The subsequent adaptation phase involves freezing the classifier $f$ as fixed task-specific anchors to prevent catastrophic forgetting, while updating the feature extractor $g$ through techniques via information-theoretic objectives~\citep{infomax} and pseudo-label refinement.
The efficacy of this process is fundamentally constrained by the cross-domain discrepancy, the inductive bias of the feature mapping $g$, and the quality of generated pseudo-labels.
Notably, while cross-domain discrepancy is an inherent property of the data domains, both feature transferability and pseudo-label quality fundamentally depend on the model's architectural design and learned representations.

\subsection{Preliminaries}
\paragraph{SSMs}
State space models (SSMs) are a class of sequence architectures that combine elements of recurrent neural networks and CNNs, with a hybrid nature that stems from their dual capability to process sequential data with recurrent-style hidden states while maintaining convolutional-like parallel computation during training. 
Specifically, SSMs define a mapping function $x(t)\xrightarrow{} y(t)$ through a hidden state $h(t) \in \mathbb{R}^N$. Formally, SSMs are governed by the continuous-time ordinary differential equation:
\begin{align}
h'(t) &= \mathbf{A}h(t) + \mathbf{B}x(t),\\
h(t)&=\overline{\mathbf{A}} h(t-1)+\overline {\mathbf B}x(t), \label{eq:state_pro}\\
y(t) &= \mathbf{C}h(t) + \mathbf{D}x(t),
\end{align}
where 
$\mathbf{A} \in \mathbb{R}^{N \times N}$, $\mathbf{B} \in \mathbb{R}^{N \times 1}$, $\mathbf{C} \in \mathbb{R}^{1 \times N}$, $\mathbf{D} \in \mathbb R^1$ are weighting matrices, and $\overline{\mathbf A},\overline{\mathbf B}$ are discrete counterparts.

Mamba~\citep{mamba} distinguishes itself from previous SSMs by dynamically conditioning its time-scale parameter $\Delta$ and state transition weights $\mathbf{B}$, $\mathbf{C}$ on the input sequence $x(t)$. This input-dependent parameterization enables Mamba to advance in contextual adaptability while maintaining computational efficiency. 

In this work, we adopt the 2D selective scan mechanism from Visual Mamba~\citep{vmamba} for vision data processing. This approach extends the sequential modeling capability of Mamba to images through a four-way scanning strategy that comprehensively traverses spatial dimensions.

\paragraph{Deep Clustering Labeling}
Deep clustering has been widely adopted in SFDA methods~\citep{SHOT,RGV,UPA,DPL} for generating pseudo-labels on target domain data. Following the established paradigm in SHOT~\citep{SHOT}, we generate pseudo-labels through class-aware feature clustering. The process begins with computing initial class centroids via soft clustering of target features:
\begin{equation}
\mu_c^{(1)} = \frac{\sum_{x_t} \delta_c(x_t)g(x_t)}{\sum_{x_t} \delta_c(x_t)},
\label{eq:centroid_init}
\end{equation}
where $\delta_c(\cdot)$ denotes the c-th class probability from the softmax operation $\delta(\cdot)$, and initial pseudo-labels are then assigned by nearest-centroid classification using the cosine similarity metric $\mathcal D_{\text{cos}}(, )$:
\begin{equation}
\hat{y}^{(1)} = \arg\max_c \mathcal D_{\text{cos}}\big(g(x_t),\mu_c^{(1)}\big).
\label{eq:label_init}
\end{equation}

Final centroids and labels are then computed using hard assignments:
\begin{equation}
\mu_c^{(2)} = \frac{\sum_{x_t:\hat{y}^{(1)}=c} g(x_t)}{|{x_t:\hat{y}^{(1)}=c}|}, \quad
\hat{y}^{(2)} = \arg\max_c \mathcal {D}_{\text{cos}}\big(g(x_t),\mu_c^{(2)}\big).
\label{eq:label_final}
\end{equation}

We simplify the pseudo-label notation by using $\hat{y}$ to represent $\hat{y}^{(2)}$ in the subsequent section.

\subsection{Framework Overview}
As illustrated in Fig.~\ref{fig:frame}, our framework adapts a pre-trained Visual Mamba (VMamba) encoder~\citep{vmamba} as the feature backbone to better overcome the domain shift encountered in source-free model transfer. We enhance the encoder by appending $L\times$\shorta{} blocks, \revise{which instill long-range, channel-wise dependencies into the feature representations during the source pre-training phase. The detailed implementation is provided in Section~\ref{sec:chvss}}. The resulting transferable representations are then classified by a task-specific classifier (MLP layer).
During the source model training phase, following~\citep{SHOT}, we employ a cross-entropy loss with label smoothing~\citep{LabelSmooth}, defined as:
\begin{align}
    \mathcal{L}_{LCE} &= -\mathbb{E}_{(x_s, y_s) \in (\mathcal{X}_s, \mathcal{Y})} \sum_{c=1}^{C} \tilde y_c\log \delta_c(\theta_s(x_s)), \\
    \ie,{} \tilde{y}_c &= (1-\alpha)\cdot\mathbb{1}_{[c=y_s]} + \alpha/C, \nonumber
\end{align}  
where $\mathbb{1}_{[c=y_s]}$ represents the indicator function, $\alpha=0.1$ denotes the smoothing parameter. 

For target adaptation, we initialize the model with the source-pre-trained parameters. The classifier is frozen to provide stable task guidance. 
Images $x_t$ of each batch first go through the patch portion layer to be patch embedding $I$, then activation maps are generated after going through the Mamba encoder, \shortb{} blocks, batch normalization (BN) layer, and ReLU activation.
To enhance the robustness of the spatial scanning process, our \shortb{} strategy identifies low-activation regions in the latent feature space $\mathbb R^D$ via GradCAM~\citep{Grad-cam}, maps them to patch embeddings as background patches, shuffles these embeddings as $\tilde I$, and reintegrates $\tilde I$ back into the framework. A consistency regularization loss $\mathcal L_{KL}$ is applied between the original and perturbed outputs; details are offered in Section~\ref{sec:scs}.
Following established SFDA practices~\citep{SHOT,NAMI,NRC}, we incorporate an information maximization objective to prevent overconfidence and encourage balanced predictions. This consists of an entropy minimization term $\mathcal{L}_{ent}$ and a diversity regularization term $\mathcal{L}_{div}$~\citep{infomax}:
\begin{align}
     \mathcal L_{ent} &=-\mathbb E_{x_t\in{\mathcal{X}}_t}\sum_{c=1}^{C}\delta_c\big(\theta_t(x_t)\big)\log\delta_c\big(\theta_t(x_t)\big),\\
    \mathcal{L}_{div} &= \sum_{c=1}^{C}\hat p_c\log(\hat p_c) = \mathcal D_{\text{KL}}\big(\hat p,\frac{1}{C}\bm{1}(C)\big) - \log C, 
\end{align}
where $\hat p = \mathbb{E}_{x_t\in {\mathcal{X}}_t}[\delta(\theta_t(x))]$ represents the mean output embedding for the whole target dataset, and $\bm{1}(C)$ is a vector of ones with dimension $C$.
Furthermore, we generate pseudo-labels $\hat y$ via deep clustering to supervise the target adaptation using a cross-entropy loss:
\begin{equation}
    \mathcal L_{CE}= -\mathbb E_{x_t\in \mathcal{X}_t} \sum_{c=1}^{C}\hat y_c\log\delta_c\big(\theta_t(x_t)\big).
    \label{loss:ce}
\end{equation}

\begin{figure}[!t]
    \centering
    \includegraphics[width=0.98\linewidth]{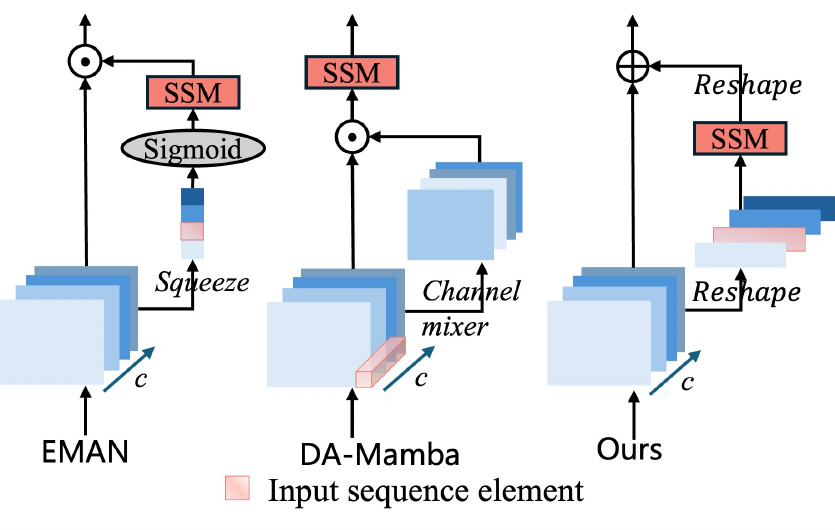}
    \caption{Comparison of channel-wise attention mechanisms (operating along the channel dimension $c$) with EMAN~\citep{EMAN} and DA-Mamba~\citep{DAMamba}, integrated into the Mamba backbone with state space modeling across different input sequence elements.}
    \label{fig:mamba_attention}
\end{figure}
\subsection{Channel-wise Visual State-Space Block (\shorta)}
\label{sec:chvss}
Traditional state space models (SSMs) primarily focus on capturing long-range spatial dependencies within sequences, often overlooking the critical role of cross-channel interactions in learning domain-invariant representations for adaptation tasks. While recent efforts like EMAN~\citep{EMAN} and DA-Mamba~\citep{DAMamba} (conceptually illustrated in Fig.~\ref{fig:mamba_attention}) have begun to explore channel-wise SSMs, they are not tailored for the source-free domain adaptation (SFDA) setting. Specifically, EMAN integrates channel attention within a complex multi-layer framework for super-resolution, and DA-Mamba necessitates concurrent access to data from both source and target domains to facilitate inter-domain channel mixing. In contrast, our SFDA scenario requires the source-trained backbone to remain fixed and operates using only unlabeled target data.

To bridge this gap, we propose the \modela{} (\shorta) block. Designed as an auxiliary neck network after the feature backbone, \shorta{} explicitly models channel-wise relationships through a novel bifurcated state propagation mechanism, effectively capturing domain-invariant frequency characteristics without altering the pre-trained backbone or requiring source data.
    
Given a target input image $x_t$ and its corresponding feature map $X_{in}=g(x_t)\in\mathbb R^{D\times H\times W}$, we first reshape the feature map into a sequence of channel vectors: $X'_{in}\in \mathbb R^{D\times(HW)}$. A bidirectional $\text{SSM}^{bi}$ is then applied along the channel dimension:
\begin{equation}
    X'_{out}=\text{SSM}^{bi}(X'_{in}) \in \mathbb R^{D\times(HW)}.
\end{equation}

$X'_{out}$ is then reshaped back into $X_{out}\in \mathbb R^{D\times H \times W}$ and combined with the original features $X_{in}$ via residual connections, preserving spatial coherence from the encoder.
This channel-wise scanning inherently captures cross-domain consistent patterns, such as edge detectors, while maintaining linear computational complexity with respect to the channel dimension $D$. The final outputs of our model are generated as:

\begin{equation}
    \theta_t(x_t) = f\Big(\text{\shorta}\big(g(x_t)\big)\Big).
\end{equation}

\begin{figure}[!t]
    \centering
    \includegraphics[width=0.9\linewidth]{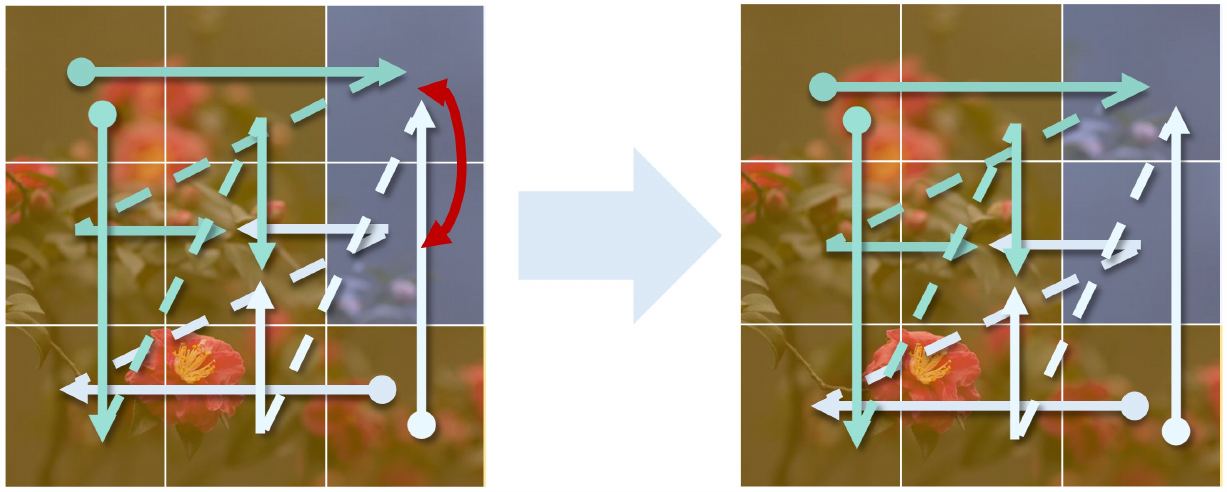}
    \caption{Comparison of the four-way 2D selective scan sequences before and after background patch shuffling in \shortb{} strategy.}
    \label{fig:scs}
\end{figure}

\subsection{Semantic-Consistent Shuffle (\shortb)}
\label{sec:scs}
The exceptional capability of Mamba models in capturing long-range dependencies stems fundamentally from their selective scan mechanism, where the evolution of the hidden state $h(t)$ depends critically on the sequential order of the input patches $x(t)$ through parameterized state transitions, as formalized in Eq.~\ref{eq:state_pro}. However, this strength becomes a vulnerability under domain shift: when contextual cues learned from the source domain are absent or altered in the target domain, the model experiences significant performance degradation, which is a manifestation of shortcut learning where spatial context serves as a brittle heuristic.

To address this issue, we propose \modelb{} (\shortb{}), a strategy that disrupts the model's reliance on spurious spatial correlations by enforcing semantic consistency between original and perturbed inputs.
The core of \shortb{} is a targeted sequence reprogramming operation. It first identifies background patches $\{P_k\}_{k=1}^N$ as the $\gamma\%$ of patches with the lowest Grad-CAM~\citep{Grad-cam} activation scores w.r.t. the pseudo-label $\hat{y}$. These patches are then shuffled via a random permutation matrix $\Pi$ to form a perturbed sequence $\widetilde{I}$:
\begin{equation}
    \widetilde P_k = P_{\Pi(k)}, \quad \widetilde I = \text{Replace}(I, \{\widetilde P_k\}).
\end{equation}
This operation deliberately disrupts the fixed contextual relationships that the model has overfitted to. As illustrated in Fig.~\ref{fig:scs}, SCS effectively modifies the scanning sequence in VMamba’s four 2D scan routes.  

We then enforce consistency between the original and perturbed sequences through a KL-divergence–based regularization loss:
\begin{equation}
    \mathcal{L}_{\text{KL}} = \mathbb{E}_{x_t \sim \mathcal{X}_t} \mathcal{D}_{\text{KL}}\big(\theta_t(I) \parallel \theta_t(\widetilde I)\big),
\end{equation}
which compels the model's hidden state dynamics to remain invariant to the order of background elements. In essence, \shortb{} trains the SSM to evolve its state primarily from robust foreground features, effectively ignoring noise introduced by spurious spatial sequences.

Furthermore, to reduce the impact of noisy pseudo-labels, we restrict the computation of $\mathcal L_{CE}$ and $\mathcal L_{KL}$ to a selected subset $\mathcal X^{sel}_t$ of the target data, obtained via feature-space neighbor perception~\citep{UPA}:
\begin{align}
\mathcal L_{CE}&=-\mathbb E_{x_t\in \mathcal{X}_t^{sel}} \sum_{c=1}^{C}\hat y_c\log\delta_c\big(\theta_t(x_t)\big), \\
\mathcal{L}_{KL} &= \mathbb E_{x_t\sim \mathcal X^{sel}_t} \mathcal D_{\text{KL}}\big(\theta_t(I)||\theta_t(\widetilde I)\big).
\end{align}
Implementation details for $\mathcal X^{sel}_t$ are provided in the \ref{sec:sel_upa}.

\subsection{Final Loss}
The overall training loss for the target model's adaptation is then defined as:
\begin{equation}
    \mathcal L_T= \mathcal{L}_{KL} +\mathcal{L}_{ent}+\mathcal{L}_{div}+\mathcal L_{CE}.
    \label{loss:all}
\end{equation}

\input{office-home}
\input{office}
\input{domainnet126}

\section{Experiments}
\label{sec:exp}
\subsection{Experiment Setups}
\subsubsection{Datasets} We evaluate our approach on four benchmarks: (1) Office~\citep{office-31}: The small-scale dataset consists of 31 classes across \textbf{A}mazon, \textbf{D}slr, and \textbf{W}ebcam domains; (2) Office-Home~\citep{office-home}: A medium-scale dataset that comprises 65 classes, distributed across four domains: \textbf{A}rt, \textbf{C}lipart, \textbf{P}roduct and \textbf{R}eal-World. (3) VisDA-C~\citep{Visda-c}: A large-scale dataset featuring 12 classes, focusing on an adaptation task from the source \textbf{S}ynthetic domain to target \textbf{R}eal domain. (4) DomainNet126~\citep{domainnet126}: a curated subset of DomainNet~\citep{DomainNet} comprising 126 object classes across four distinct domains: \textbf{C}lipart, \textbf{P}ainting, \textbf{R}eal, and \textbf{S}ketch.

\subsubsection{Implementation Details}
We adopt Vmamba-S (Small) and Vmamba-T (Tiny) as our backbone network for benchmark evaluations. The architecture modifies the original Vmamba by replacing the final classifier with our proposed \shorta{} block ($L$=2), followed by batch normalization and ReLU activation layers, and concluding with a task-specific classifier head.
The source model training is conducted for 100, 50, 10, and 30 epochs on Office, Office-Home, VisDA-C, and DomainNet126 datasets, respectively. We set the initial learning rate $lr$ to $3\times 10^{-4}$ for Office, Office-Home and DomainNet126, while using $3\times 10^{-5}$ for VisDA-C. 
During target adaptation, we set $lr'=5\times 10^{-5}$ for Office, Office-Home, and $lr'=0.1lr$ for VisDA-C and DomainNet126. The classifier layer remains fixed during this phase, with the feature backbone using a reduced learning rate of $0.1lr'$. All training utilizes AdamW optimizer with a weight decay of 5e-2 and cosine annealing learning rate scheduling. Experiments were executed with a batch size of 32 on a NVIDIA RTX A100 GPU, running for 15 epochs except DomainNet126, which required only 5 epochs due to observed faster convergence. The background ratio $\gamma$\% is set to 20\% through ablations. \revise{All results are averaged over three runs with three random seeds $\{0, 1, 2\}$.} 

\subsection{Main Results} The main results are summarized in Table~\ref{tab:office_home}--Table~\ref{tab:domainnet126}, comparing \shortname{} against existing methods based on CNN, ViT, and Mamba backbones.  Additional efficiency analysis of \shortname{} is provided in Section~\ref{compute_cost}, demonstrating \shortname{}'s advantages in computational performance.

\paragraph{\textbf{Office-Home}}
Our source-only models Source-T (72.3\%) demonstrate a stronger baseline compared to the source-only baselines including ResNet-50 (59.6\%) and DeiT-S (69.8\%), highlighting the effectiveness of our initial architectural enhancement. After adaptation, \shortname{}-S sets a new SOTA with 81.7\% average accuracy, surpassing the previous best ViT-based method, C-SFTrans (80.6\%), by 1.1\% and previous Mamba-based DATMamba (80.9\%) by 0.8\%. Notably, our method achieves the best performance in 8 out of the 12 transfer tasks.

\paragraph{\textbf{VisDA-C}}
As shown in the last column in Table~\ref{tab:office_home}, on the challenging VisDA-C benchmark, \shortname{}-S achieves 89.3\% accuracy, exceeding DATMamba and surpassing C-SFTrans (88.3\%) by 1.0\%, while also outperforming the non-source-free CDTrans-B (88.5\%). The compact \shortname{}-T variant attains 88.5\% accuracy, competitive with larger ViT-based models while maintaining superior efficiency. The substantial improvement from the source-only model with \shorta{} block (65.6\%) to the complete SfMamba framework (89.3\%) validates the crucial role of our \shortb{} strategy in handling significant domain shifts.

\paragraph{\textbf{Office}}
As indicated in Table~\ref{tab:office_uda}, our source-only models Source-T (88.4\%) and Source-S (89.5\%) outperform ResNet-50 (77.2\%), DeiT-S (86.7\%), and DeiT-B (88.8\%). After adaptation, \shortname{}-S achieves state-of-the-art performance with 93.3\% average accuracy, demonstrating particularly strong improvements on the D$\xrightarrow{}$A and W$\xrightarrow{}$A tasks.

\paragraph{\textbf{DomainNet-126}}
In the Table~\ref{tab:domainnet126}, \shortname{}-S achieves the highest average accuracy of 77.9\%, outperforming the previous best source-free method SHOT-S by 1.4\% and attaining top performance in six of the seven transfer tasks. Notably, \shortname{}-S surpasses the non-source-free CDTrans-B (75.8\%) by 2.1\%, despite the latter utilizing a larger DeiT-B backbone and requiring source data access. Furthermore, our method exceeds the best CNN-based approach RGV (73.2\%) by 4.7\%, demonstrating its substantial advantages across different architectural paradigms.

\input{abl_module}
\subsection{Ablation Study}
The following ablation studies employ the VMamba-S backbone unless otherwise specified.
\subsubsection{Ablation for Proposed Modules}
We evaluate the proposed modules on Office-Home and VisDA-C datasets, with quantitative results presented in Table~\ref{tab:abl_module}. For the target model, the \shorta{} block enables long-range dependency modeling, yielding gains of +3.4\% ($\text{Baseline}_T$ $\xrightarrow{}$ $w/$ \shorta) and +2.7\% ($w/$ SCS $\xrightarrow{}$ \shortname) on Office-Home dataset after adaptation, with even greater improvements of +4.5\% and +6.1\% on VisDA-C. The \shortb{} module further enhances performance, delivering improvements of +1.2\% ($\text{Baseline}_T$ $\xrightarrow{}$ $w/$ \shortb) and +0.5\% ($w/$ \shorta{} $\xrightarrow{}$ \shortname) on Office-Home, along with corresponding gains of +2.3\% and +3.9\% on VisDA-C.
The comprehensive adaptation capability is particularly evident in the end-to-end improvements from source to target model, where the Baseline shows gains of +3.4\% and +13.6\%, while the \shortname{} achieves better performance with improvements of +6.4\% and +23.7\% compared to Source-S. 
These consistent improvements across both datasets validate the complementary benefits of integrating \shorta{} and \shortb{} for SFDA.

\begin{figure}[!t]
    \centering
    \subfloat[\srevise{A$\xrightarrow{}$C}]{\includegraphics[width=0.48\linewidth]{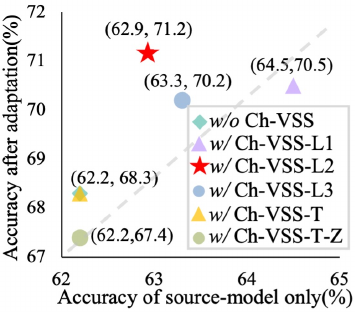}
    \label{fig:abl_vss_layer}
    }
    \subfloat[]{   
    \includegraphics[width=0.48\linewidth]{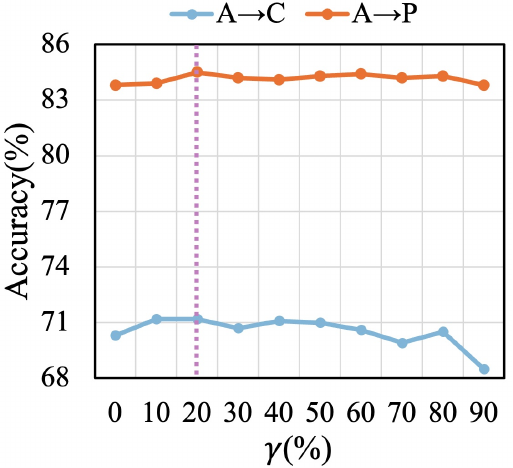}
    \label{fig:abl_con_ratio}
    }
    \caption{Hyperparameter analysis: (a) \srevise{Accuracy of the source-model-only and adapted target model on the }Office-Home A$\xrightarrow{}$C task across different \shorta-$\textbf{L}$ayer$\textbf{n}$ configurations; (b) Impact of background ratio $\gamma$ on target accuracy for A$\xrightarrow{}$C and A$\xrightarrow{}$P tasks in Office-Home dataset.} 
    \label{fig:abl_gamma}
\end{figure}

\subsubsection{Hyperparameter Analysis}
This section examines two critical hyperparameters: the number of stacked \shorta{} blocks and the background patch ratio $\gamma$ used in \shortb{}.
We denote the configuration of \shorta{} blocks as \shorta-L$n$, where $n$ indicates the number of blocks. 
Fig.~\ref{fig:abl_vss_layer} reports the accuracy of both the source and adapted target models on the target domain. The results show that integrating \shorta{} blocks consistently enhances target accuracy, with the \shorta-L2 configuration yielding the optimal performance on target domain.
The impact of $\gamma$ (\%) is evaluated in Fig.~\ref{fig:abl_con_ratio}. The highest accuracy occurs at $\gamma=20$, with performance remaining stable for smaller values but degrading for larger $\gamma$. This degradation aligns with the expectation that excessive shuffling of background patches disrupts structurally important image content, thereby impairing the model's ability to learn coherent visual representations. 

\srevise{
\subsubsection{Necessary Pre-training for \shorta}
We ablate the optimal stage for applying Ch-VSS. The results are presented in Fig.~\ref{fig:abl_vss_layer}. Applying it only during target adaptation ($w/$ Ch-VSS-T) yields no improvement over the baseline. Even when we append zero-initialized convolutional layers~\citep{contronet} specifically to the \shorta{} blocks at the start of the target adaptation phase (denoted as $w/$ Ch-VSS-T-Z) to mitigate feature disruption, performance still does not improve. These results confirm that pre-training the Ch-VSS module on the source domain is essential. It allows the module to learn a foundational, domain-invariant feature reorganization policy, serving as a necessary precursor for effective target adaptation. }

\input{abl_attnv2}
\input{abl_backbone}
\input{abl_scs}
\begin{figure}[!t]
    \centering
\includegraphics[width=0.9\linewidth]{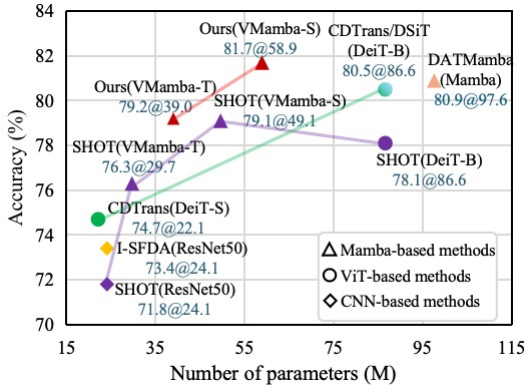}
   \caption{Comparison of accuracy versus parameter efficiency across backbones on Office-Home. Individual methods are distinguished by unique colors. Our \shortname{} achieves a favorable trade-off.}
    \label{fig:acc_pam}
\end{figure}

\subsubsection{Effectiveness of \shorta}
To evaluate the efficacy of \shorta{}, we conduct ablation studies with other representative channel-wise attention mechanisms: self-attention (SA)~\citep{vit} used in transformer~\citep{vit}, channel Mamba (CM)~\citep{EMAN}, squeeze-and-excitation (SE)~\citep{SE} along channel dimension, efficient channel attention (ECA)~\citep{eca}, and convolution block attention (CBAM)~\citep{CBAM}.  \srevise{To critically assess the sufficiency of the bidirectional scan within Ch-VSS, we designed a controlled ablation: a Group-wise Channel Scanning (Ch-Group) strategy. This more complex variant forcibly creates a shuffled channel scanning sequence by reshaping the flattened 1D channel features into a 2D grid of width $d$ and applying full bidirectional cross-scanning across both rows and columns (visualized in APPENDIX Fig.~\ref{fig:cross2d}).}
\srevise{As shown in Table~\ref{tab:abl_attn}, while the alternative attention mechanisms improve target domain accuracy by at most 0.5\% on average, our \shorta{} module yields a significant +2.5\% gain, with improvements observed across all transfer tasks.
Furthermore, the Ch-Group variant fails on several transfer tasks, \eg, A$\xrightarrow{}$C and A$\xrightarrow{}$P. This result constructively proves that the channel dependencies necessary for effective adaptation are already being captured efficiently by the intrinsic bidirectional scan in our Ch-VSS design, making more complex, forced permutations unnecessary.}

\srevise{To further validate the generalizability, we integrated Ch-VSS into diverse backbones: Agent-DeiT~\citep{agent_attention}, DeiT~\citep{DeiT}, and ConvNeXt~\citep{convnext}. Table~\ref{tab:abl_backbone} confirms that Ch-VSS consistently improves target domain performance across all architectures while maintaining parameter efficiency. The best results are achieved with the native VMamba-S backbone, underscoring a synergistic design within the state-space framework.}

\subsubsection{Effectiveness of \shortb}
We conduct an ablation study to evaluate the background identification strategy in \shortb{}, with results presented in Table~\ref{tab:abl_scs}. Three configurations are compared: no shuffling (None), random shuffling (Random), and our selective approach (\shortb). The empirical results demonstrate two key findings: (1) disrupting spatial coherence through patch shuffling significantly improves performance (None vs. Random), and (2) selectively identifying background patches for shuffling yields additional gains (Random vs. SCS). Notably, on the challenging VisDA-C (S$\xrightarrow{}$R) task, SCS achieves a significant accuracy of 89.3\%, substantially outperforming None (85.4\%) and Random (88.5\%).
These results confirm that our background disruption strategy effectively enhances model robustness by maintaining focus on semantically relevant foreground content during state-space modeling.

\input{flops}
\begin{figure*}[!th]
    \centering
    \includegraphics[width=0.85\linewidth]{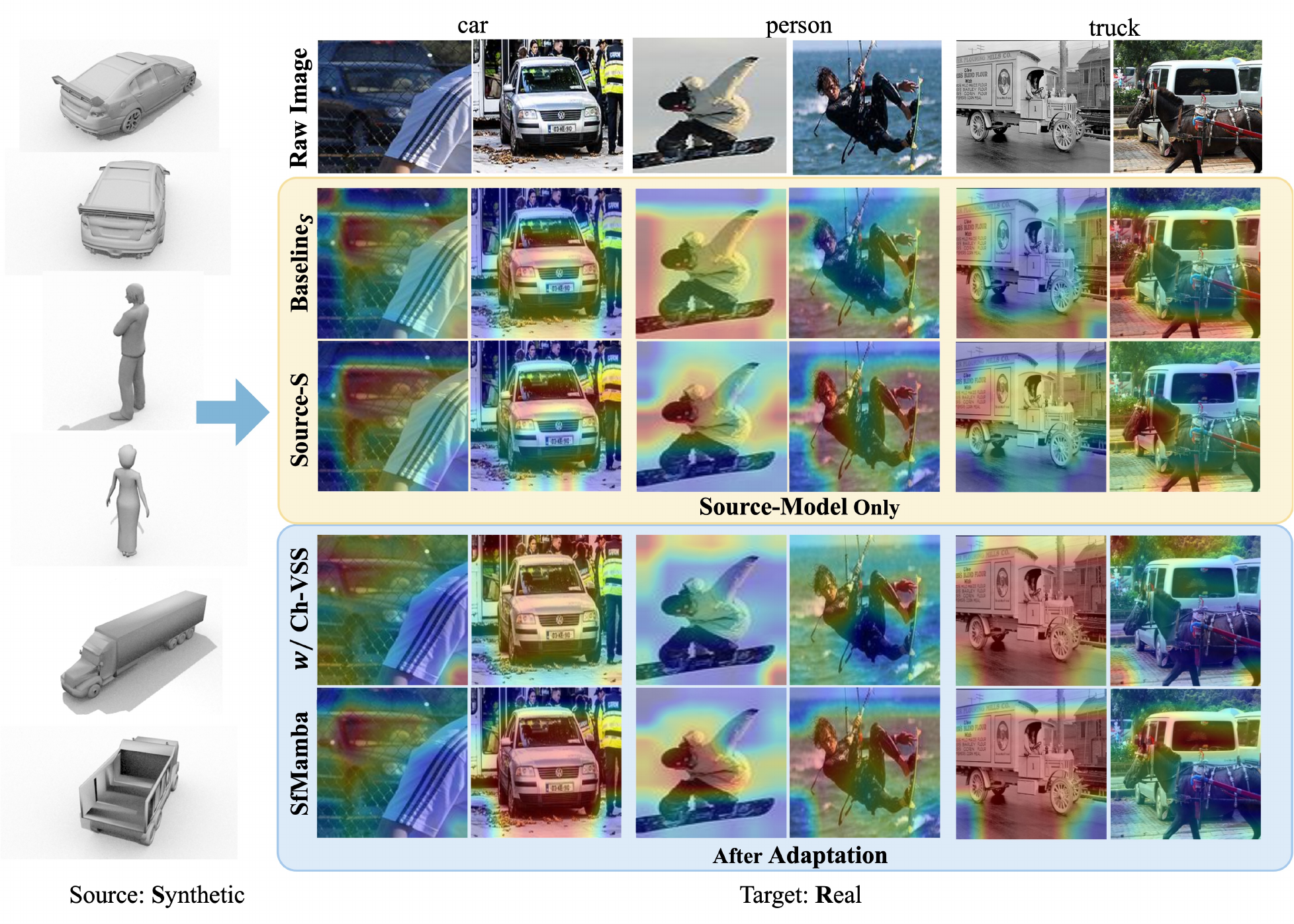}
    \caption{Grad-CAM visualization results from the module ablation study on the target \textbf{R}eal domain in VisDA-C dataset. Each row corresponds to an experimental setting in Table~\ref{tab:abl_module}, comparing the source-only model and the adapted target model for ablations of \shorta{} and \shortb{}.}
    \label{fig:abl_cam}
\end{figure*}
\begin{figure*}[!th]
    \centering
    \includegraphics[width=0.85\linewidth]{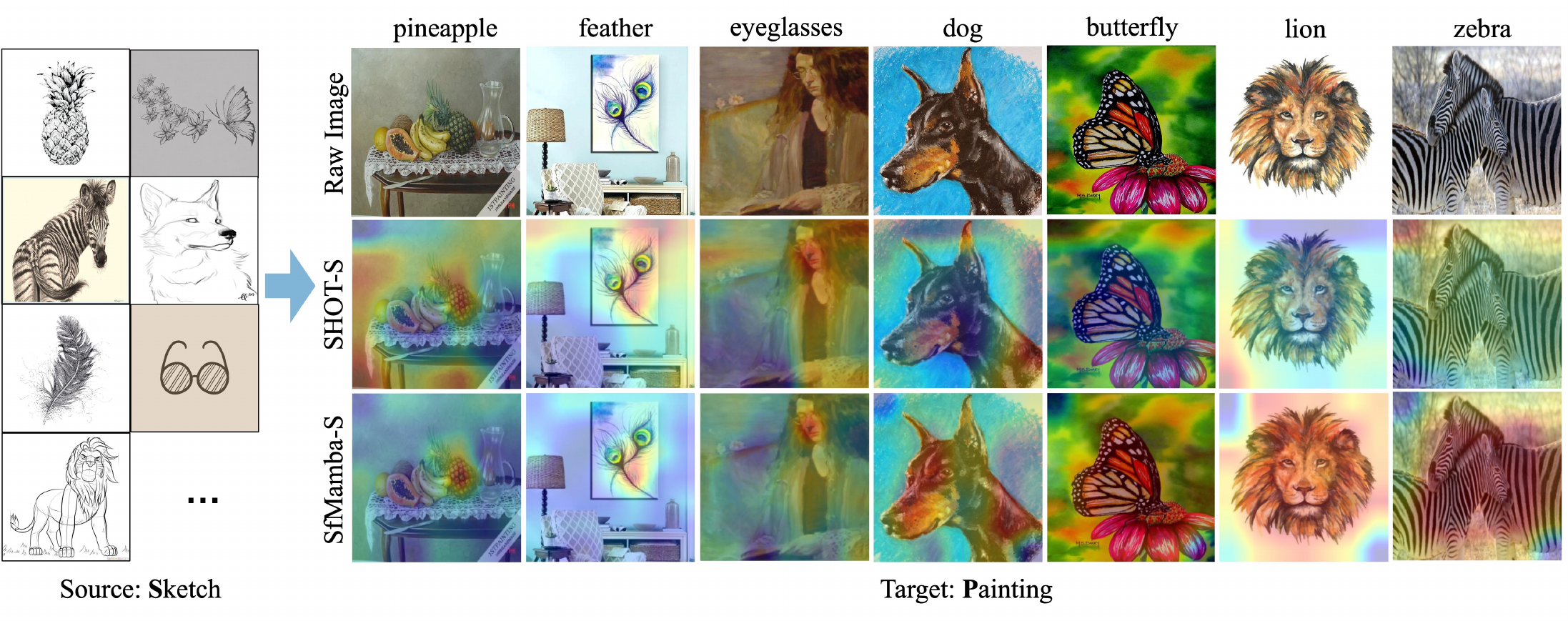}
    \caption{Grad-CAM visualizations for the S$\to$P transfer task on DomainNet-126, comparing \shortname-S and SHOT-S, both based on the VMamba-S backbone.}
    \label{fig:gradcam}
\end{figure*}

\subsubsection{Computational Cost}
\label{compute_cost}

Fig.~\ref{fig:acc_pam} presents the accuracy-parameter trade-off comparison to CNN, ViT, and Mamba alternatives, where \shortname{} achieves accuracy-parameter efficiency. Our VMamba-S variant attains 81.7\% accuracy with only 58.9M parameters, outperforming DeiT-B based CDTrans/DeSiT (80.5\%with 86.6M). Similarly, VMamba-T achieves 79.2\% accuracy with 39.0M parameters, surpassing comparable ViT and CNN-based methods. 
Further computational analysis in Table~\ref{tab:flops} reinforces this efficiency. Compared to the ResNet101-based I-SFDA method, our VMamba-T implementation reduces the parameter count by 4M and computational cost by 2.4 GFLOPs. The efficiency gains are even more pronounced against DeiT-B based CDTrans~\citep{CDTrans}, where our VMamba-S variant achieves superior accuracy while using 32.0\% fewer parameters (27.7M), 47.7\% fewer FLOPs (8.4G), and a 25.9\% higher throughput. Moreover, against the Mamba-based DATMamba, \shortname{} reduces parameters and FLOPs by 38.7M (39.7\%) and 5.8G (38.7\%), respectively. These results underscore the compelling computation-accuracy trade-off of our method.

\subsection{Visualization Analysis}
Fig.~\ref{fig:abl_cam} presents an ablation study on the VisDA-C S$\to$R transfer task. Our analysis reveals that while the baseline source model exhibits only rudimentary object awareness on target data, augmenting it with our \shorta{} module improves generalization, enabling more accurate identification of task-relevant regions even before adaptation. After full adaptation, the heatmaps of \shortname{} become significantly sharper and more concentrated on foreground objects. Critically, conventional adaptation strategies are observed to degrade performance in some cases, as evidenced by the loss of object focus in person class examples, which is a result worse than the source-only model. \shortname{} overcomes this limitation through the \shortb{} strategy, which reduces noise propagation during sequence scanning in SSMs and stabilizes the learning process.

Further insights are provided in Fig.~\ref{fig:gradcam} for the S$\to$P transfer task on DomainNet-126. When compared with SHOT-S implemented on the same VMamba-S backbone, \shortname{} maintains concentrated attention on class-discriminative regions during domain transfer. In contrast, SHOT exhibits dispersed activation patterns that incorporate more domain-specific noise. This comparative analysis confirms that \shortname{} effectively preserves semantic coherence under domain shift.

\srevise{
\section{Limitations and Future works}
\label{sec:limit}
This work presents a novel Mamba-based method for SFDA, introducing channel-wise and spatial-wise scanning strategies to learn domain-invariant features. However, it faces several limitations that suggest promising avenues for future research. First, as indicated by the ablation studies in Table~\ref{tab:abl_attn}, a degree of overfitting to source-domain characteristics persists when adapting from the highly stylized synthetic domain Clipart. The simplified textures and strong intra-domain consistency of these sources pose a distinct challenge (as visualized in APPENDIX Fig.~\ref{fig:fail_domains}), revealing a boundary condition of our current approach. Second, while our contributions are empirically robust, they remain primarily so. A formal theoretical analysis explaining why and how state-space models (SSMs) are particularly effective at capturing domain-invariant representations is an important open question that would deepen the understanding of this paradigm.}

\srevise{Future work could address these challenges by: (1) enhancing robustness to extreme domain shifts, potentially through dynamic style augmentation or adaptive scanning schedules; and (2) pursuing the aforementioned theoretical grounding while exploring integration with emerging SSM architectures and more advanced scanning mechanisms to push the boundaries of efficiency and generalization.}

\section{Conclusion}
\label{sec:conclusion}
This paper presents \shortname{}, a novel and efficient selective state space framework for source-free domain adaptation. By incorporating \modela{} block for domain-invariant dependency modeling and \modelb{} for noise-robust spatial adaptation, \shortname{} effectively establishes long-range dependencies critical for source-free domain transfer. Comprehensive experiments on four standard benchmarks demonstrate the framework's efficiency and robustness in SFDA.
\section*{CRediT authorship contribution statement}
\textbf{Xi Chen}: Conceptualization, Methodology, Writing – original draft. 
\textbf{Hongxun Yao}: Conceptualization, Funding acquisition, Methodology, Writing – review and editing. 
\textbf{Sicheng Zhao}: Methodology, Supervision, Formal analysis,  Writing – review and editing. 
\textbf{Jiankun Zhu}: Methodology, Formal analysis, Validation. 
\textbf{Jing Jiang}: Methodology, Formal analysis, Validation. 
\textbf{Kui Jiang}: Methodology, Supervision, Formal analysis, Writing – review and editing. 

\section*{Declaration of competing interest}  
The authors declare that they have no known competing financial interests or personal relationships that could have appeared to influence the work reported in this paper.

\section*{Acknowledgements}
This work was supported by the National Science Foundation of China under Grant 62476069.

\appendix
\section{More Target Adaptation Details}  
\subsection{$\mathcal X_t^{sel}$ Generation}
\label{sec:sel_upa}
follows UPA filtering mechanism~\citep{UPA}, which mitigates pseudo-label noise through feature-space neighborhood validation. For each target sample $x_t$, UPA identifies its $K$-nearest neighbors $\mathcal{N}_t$ using cosine similarity in the embedding space $g(\cdot)$. The class posterior distribution $\hat p_t \in \mathbb{R}^D$ is computed via distance-weighted voting:
\begin{equation}
\hat p_t=\frac{1}{|\mathcal N_t|} \sum_{x_k\in \mathcal{N}t} \mathcal D_{\text{cos}}\big(g(x_t),g(x_k)\big) \bm{e}_{\hat y_k},
\label{eq:cls_prob}
\end{equation}
where $\bm{e}_{\hat y_k}$ is the one-hot vector of the initial pseudo-label $\hat y_k$ obtained via deep clustering. The refined pseudo-label $\hat{y}'_t = \arg\max_c \hat{p}_t$ is then iteratively fed back into Eq.~\ref{eq:cls_prob} as new pseudo-labels for further refinement of $\hat p_t$ and $
\hat y'_t$. The final confidence score $q_t$ is computed by considering only consensus neighbors sharing the refined label $\hat y_t'$:

\begin{equation}
q_t = \frac{1}{|\mathcal{N}_t^\text{cons}|} \sum_{\substack{x_k \in \mathcal{N}_t^\text{cons}}} \mathcal{D}_{\text{cos}}(g(x_t), g(x_k)),
\end{equation}
where $\mathcal{N}_t^\text{cons} = \{x_k \in \mathcal{N}_t | \hat{y}'_k = \hat{y}_t\}$. All target samples are then ranked within each class according to their consensus confidence scores $q_t$, with the top $\beta=60\%$ of each class constituting the selected set $\mathcal{X}_t^{sel}$ of high-confidence pseudo-labeled target samples.

\subsection{Ablations on $\mathcal{X}_{t}^{sel}$}
We conduct a comprehensive ablation study to examine the effect of sample filtering (producing $\mathcal{X}_{t}^{sel}$) on target domain accuracy, as presented in Table~\ref{tab:suppl_abl_module}. The results demonstrate that while the filtering mechanism ($w/$ Filter) provides some adaptation benefits, it fails to surpass the performance of \shortname{}. This finding suggests that conventional noise-filtering pseudo-label learning approaches~\citep{UPA} in SFDA are insufficient to optimize Mamba's performance, highlighting the need for our proposed \shortb.
\input{suppl_abl_module}

\srevise{\section{More Visualized Illustration}}
\begin{figure}[!th]
    \centering
    \includegraphics[width=\linewidth]{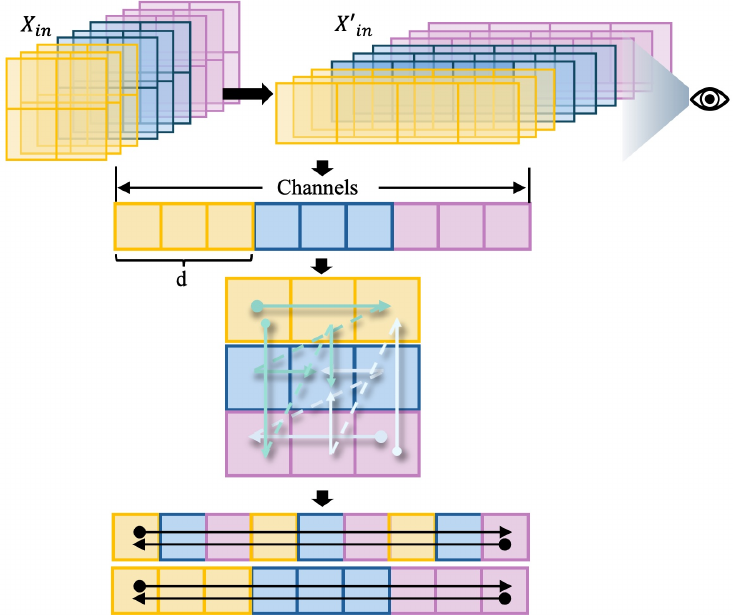}
    \caption{\srevise{Schematic of Ch-Group: a channel permutation mechanism that groups channels into segments of size $d$ and sequentially interleaves them to disrupt sequential connectivity patterns.}}
    \label{fig:cross2d}
\end{figure}
\begin{figure}[!th]
    \centering
\includegraphics[width=\linewidth]{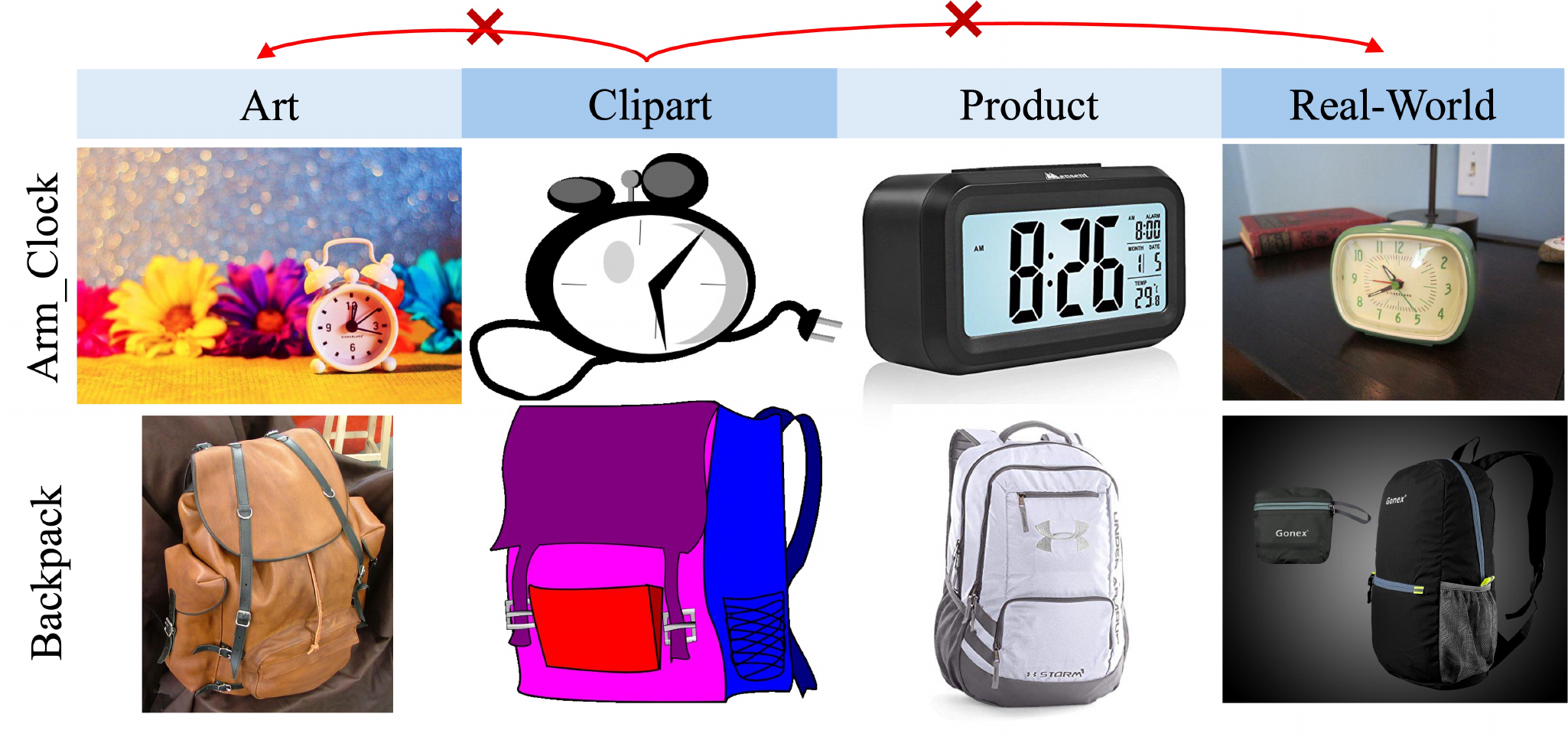}
\caption{\srevise{Visual examples from the Office-Home dataset, illustrating the challenging adaptation tasks from the source Clipart domain to target domains.}}
    \label{fig:fail_domains}
\end{figure}

\srevise{\subsection{Schematic of Channel Attention Case Ch-Group}
The schematic of the Ch-Group mechanism is provided in Fig.~\ref{fig:cross2d}, where we set $d=6$. The resulting shuffled scanning sequences are subsequently fed into the SSM block for processing.}

\srevise{
\subsection{Limitations on Source Domain Learning}
Fig.~\ref{fig:fail_domains} illustrates a key limitation: the risk of overfitting when models are pre-trained on highly stylized source domains such as Clipart. The figure juxtaposes examples from this synthetic source with its real-world target domains in Office-Home, highlighting the substantial visual disparity that complicates adaptation.}

\bibliographystyle{elsarticle-num-names}
\bibliography{Reference_v2}

\end{document}

%% file: office-home.tex
\begin{table*}[!th]
\centering
\setlength{\tabcolsep}{5pt}
\caption{Accuracy(\%) on Office-Home and VisDA benchmarks. The ``SF'' column indicates whether a method is source-free ($\checkmark$) or not (\ding{55}). The compared CNN-based methods use ResNet50 for Office-Home and ResNet101 for VisDA-C. \dag{}: Results from official code reproduction. Best and second-best results are highlighted in \textbf{bold} and \underline{underline}, respectively.}
\small
\resizebox{.98\textwidth}{!}{
\begin{tabular}{l|c|*{13}{c}|c}
   \toprule
    \multirow{2}{*}{Method} &\multirow{2}{*}{SF}&\multicolumn{13}{c}{Office-Home}&\multicolumn{1}{|c}{VisDA-C}\\
    \cmidrule{3-16}
    &&A$\xrightarrow{}$C& A$\xrightarrow{}$P& A$\xrightarrow{}$R& C$\xrightarrow{}$A& C$\xrightarrow{}$P& C$\xrightarrow{}$R &P$\xrightarrow{}$A &P$\xrightarrow{}$C& P$\xrightarrow{}$R& R$\xrightarrow{}$A& R$\xrightarrow{}$C& R$\xrightarrow{}$P & Avg&S$\xrightarrow{}$R\\
    \midrule
    \multicolumn{16}{c}{\textbf{CNN-based Methods}}\\
    \midrule
    ResNet~\citep{Resnet}& \multirow{1}*{\checkmark}&44.0& 65.9& 74.0 &52.1& 61.0 &65.4 &52.2& 40.9 &72.9& 64.1 &45.3 &77.9& 59.6&47.8 \\
   SHOT~\citep{SHOT}&\checkmark& 57.1 &78.1& 81.5&68.0 &78.2& 78.1 &67.4& 54.9 &82.2& 73.3 &58.8& 84.3 &71.8&82.9\\
    NRC~\citep{NRC} &\checkmark&57.7 &80.3& 82.0& 68.1 &79.8& 78.6 &65.3& 56.4& 83.0 &71.0& 58.6& 85.6& 72.2&85.9\\
    CoWA-JMDS~\citep{CoWA-JMDS}&\checkmark&56.9 &78.4& 81.0& 69.1 &80.0 &79.9& 67.7& 57.2& 82.4& 72.8 &60.5 &84.5& 72.5&86.9\\
NAMI~\citep{NAMI}&\checkmark&58.3&80.9&82.9&68.5&80.5&79.6&66.0&57.0&83.7&71.9&59.9&85.0&72.9&88.9\\
    UPA~\citep{UPA}&\checkmark&57.3&79.5& 82.4& 68.2& 81.0&80.2&69.3& 55.4& 82.3& 75.3&59.2&85.5& 73.0&88.7\\
    TPWA~\citep{TPWA}&\checkmark&58.8& 80.3& 82.4& 69.5 &80.6& 78.8 &68.4& 55.8 &82.0 &74.7& 59.2 &85.5 &73.0&87.2\\
    
    SHOT+DPC~\citep{DPC}&\checkmark &59.2 &79.8 &82.6& 68.9& 79.7& 79.5& 68.6& 56.5 &82.9& 73.9& 61.2 &85.4 &73.2&84.3\\
    I-SFDA~\citep{I-SFDA}&\checkmark&60.7& 78.9& 82.0 &69.9 &79.5 &79.7 &67.1& 58.8& 82.3 &74.2& 61.3& 86.4 &73.4&88.4\\
    
    C-SFDA~\citep{C-SFDA}&\checkmark&60.3& 80.2& 82.9& 69.3& 80.1& 78.8 &67.3& 58.1& 83.4& 73.6 &61.3& 86.3 &73.5&87.8\\
    TPDS~\citep{TPDS}&\checkmark&59.3& 80.3& 82.1& 70.6& 79.4& 80.9 &69.8& 56.8& 82.1& 74.5 &61.2 &85.3 &73.5&87.6\\
    DPL~\citep{DPL}&\checkmark& 60.9 &80.8 &82.3& 69.9 &79.5& 80.0& 70.0 &57.0& 82.7& 75.7& 63.0& 86.0 &74.0 & 87.8\\
    RGV~\citep{RGV}&\checkmark&61.2& 80.9 &82.7 &69.3 &81.2& 81.4 &68.1& 58.8& 83.4 &74.6& 62.4 &85.7 &74.1&\underline{89.1}\\
    DAPL~\citep{DAPL}&\ding{55} & 54.1& 84.3& 84.8& 74.4 &83.7 &85.0 &74.5 &54.6& 84.8& 75.2& 54.7& 83.8 &74.5&86.9\\
    \midrule
    \multicolumn{16}{c}{\textbf{ViT-based Methods}}\\
    \midrule
    DeiT-S~\citep{DeiT}&\checkmark&55.6& 73.0 &79.4& 70.6& 72.9 &76.3 &67.5& 51.0& 81.0& 74.5& 53.2& 82.7& 69.8&- \\
    DeiT-B~\citep{DeiT}&\checkmark&61.8& 79.5& 84.3& 75.4& 78.8& 81.2& 72.8& 55.7& 84.4 &78.3& 59.3& 86.0& 74.8&67.1 \\
    CDTrans-S~\citep{CDTrans}&\ding{55}&60.6 &79.5 &82.4 &75.6 &81.0 &82.3& 72.5 &56.7 &84.4& 77.0& 59.1& 85.5& 74.7&-\\
    SHOT-B~\citep{SHOT}&\checkmark&67.1 &83.5 &85.5& 76.6& 83.4 &83.7&76.3 &65.3& 85.3 &80.4& 66.7& 83.4& 78.1&83.5\\
    CDTrans-B~\citep{CDTrans}&\ding{55}&68.8& \underline{85.0} &86.9& \underline{81.5} &\underline{87.1}& \underline{87.3} &\underline{79.6} &63.3 &\underline{88.2} &82.0 &66.0 &\underline{90.6}& 80.5&88.4\\
    DSiT-B~\citep{Dsit} &\checkmark&69.2 &83.5& 87.3& 80.7 &86.1& 86.2 &77.9& \underline{67.9}& 86.6 &{82.4} &68.3& 89.8& 80.5&87.6 \\
    C-SFTrans~\citep{C-SFTrans}&\checkmark&\underline{70.3} &83.9& 87.3& 80.2& 86.9& 86.1& 78.9 &65.0& 87.7 &\underline{82.6}& 67.9& 90.2 &{80.6}&88.3\\
    \midrule
    \multicolumn{16}{c}{\textbf{Mamba-based Methods}}\\
    \midrule
    Source-T&\checkmark&59.0& 76.8& 83.0&69.8&76.5& 78.2&68.9& 54.6& 82.4 &74.8& 58.3& 85.4&72.3&61.5 \\
    Source-S&\checkmark&62.9 & 79.5& 84.7&74.4&80.2& 81.2& 71.9& 56.9& 84.4& 78.4& 61.9& 86.6&75.3&65.6\\
    SHOT-T\dag~\citep{SHOT}&\checkmark&64.9&80.1&83.6&75.4&81.4&81.3&71.7&60.7&84.9&77.4&66.1&87.7&76.3&81.9\\
    SHOT-S\dag~\citep{SHOT}&\checkmark&67.3&81.5&85.6&79.9&85.3&83.8&76.1&66.4&85.3&80.1&69.0&88.7&79.1&84.6\\
    \shortname-T&\checkmark&67.5& 80.9 &87.4 &77.6& 83.4&86.4&76.2 &66.1& 86.9& 78.7& \underline{69.6}& 89.8&79.2&88.5\\
    DATMamba~\citep{DATMamba}&\checkmark&67.1& \textbf{86.3} &\underline{87.8}& \textbf{81.7}& \textbf{87.3}& 85.4 &\textbf{80.2}& 66.3 &87.1 &\textbf{83.2}& 67.6 &90.5& \underline{80.9} & 88.9 \\
    \shortname-S&\checkmark&\textbf{71.2}&84.5&\textbf{88.4}&\textbf{81.7}&85.9&\textbf{88.0}&{79.0}&\textbf{69.3}&\textbf{88.5}&81.1&\textbf{72.1}&\textbf{90.7}&\textbf{81.7}&\textbf{89.3}\\
    \bottomrule
\end{tabular}}
\label{tab:office_home}
\end{table*}

%% file: office.tex
\begin{table}[!th]
    \small
    \caption{Accuracy(\%) on the Office dataset. The ``SF'' column indicates whether a method is source-free ($\checkmark$) or not (\ding{55}). \dag{}: Results from official code reproduction. Best and second-best results are highlighted in \textbf{bold} and \underline{underline}, respectively.}
    \centering
\setlength{\tabcolsep}{0.5pt}
\resizebox{\linewidth}{!}{
    \begin{tabular}{l|c|*{7}{c}}
   \toprule
     Method & SF&A$\xrightarrow{}$D& A$\xrightarrow{}$W & D$\xrightarrow{}$A &D$\xrightarrow{}$W & W$\xrightarrow{}$A& W$\xrightarrow{}$D & Avg\\
    \midrule
    \multicolumn{9}{c}{\textbf{CNN-based Methods}}\\
    \midrule
    ResNet50~\citep{Resnet} &\checkmark&79.3&72.6& 60.0&93.7&61.6&96.0& 77.2\\
    SHOT~\citep{SHOT}&\checkmark&94.0& 90.1 &74.7 &98.4& 74.3 &\underline{99.9} &88.6 \\
    NRC~\citep{NRC}&\checkmark&96.0& 90.8& 75.3& 99.0& 75.0& \textbf{100.0}& 89.4\\
    SHOT+DPC~\citep{DPC}&\checkmark & 95.9& 92.6 &75.4 &98.6 &76.2 &\textbf{100.0} &89.8\\
    UPA~\citep{UPA}&\checkmark& 96.2& 93.7& 75.0& 98.4& 76.6& 99.8& 89.9\\
   TPWA~\citep{TPWA}&\checkmark&96.2 &95.2 &75.8 &98.8& 75.6 &99.2 &90.1\\
    DIPE~\citep{DIPE}&\checkmark&96.6 &93.1& 75.5 &98.4 &77.2& 99.6& 90.1\\
    TPDS~\citep{TPDS}&\checkmark & \underline{97.1}  &94.5  &75.7 & 98.7 & 75.5 & 99.8 &90.2\\
    CoWA-JMDS~\citep{CoWA-JMDS}&\checkmark&94.4 &95.2& 76.2& 98.5& 77.6 &99.8 &90.3\\
    C-SFDA~\citep{C-SFDA}&\checkmark&96.2& 93.9& 77.3& 98.8 &77.9& 99.7 &90.5\\
    NAMI~\citep{NAMI}&\checkmark&96.6&93.5&78.8&99.0&76.2&\textbf{100.0}&90.7\\
   DPL~\citep{DPL}&\checkmark& 97.0& 93.5& 77.0 &98.6& 78.0 &\textbf{100.0}& 90.7\\
    \midrule
    \multicolumn{9}{c}{\textbf{ViT-based Methods}}\\
    \midrule
    DeiT-S~\citep{DeiT}&\checkmark&87.6& 86.9& 74.9& 97.7& 73.5& 99.6& 86.7 \\
    DeiT-B~\citep{DeiT}&\checkmark&90.8 &90.4 &76.8 &98.2 &76.4 &\textbf{100.0} &88.8 \\
    CDTrans-S~\citep{CDTrans}&\ding{55}&94.6& 93.5 &78.4 &98.2& 78.0 &99.6&90.4\\
    
    SHOT-B~\citep{SHOT}&\checkmark&95.3 &94.3& 79.4 &99.0& 80.2 &\textbf{100.0}& 91.4\\
    C-SFTrans~\citep{C-SFTrans}&\checkmark&-&-&-&-&-&-&92.3\\
    CDTrans-B~\citep{CDTrans}&\ding{55}&97.0& \underline{96.7}& 81.1 &99.0& \underline{81.9}& \textbf{100.0}& 92.6\\
    DSiT-B~\citep{Dsit}&\checkmark& \textbf{98.0}&\textbf{97.2}&81.7&\underline{99.1}&81.8&\textbf{100.0}&\underline{93.0}\\
    \midrule
    \multicolumn{9}{c}{\textbf{Mamba-based Methods}}\\
    \midrule
    Source-T&\checkmark&90.8&87.2&77.5&98.7&76.3&\textbf{100.0}&88.4\\
    Source-S&\checkmark&91.0&89.9& 78.8& 98.5&78.9& \textbf{100.0}&89.5\\
    SHOT-T\dag~\citep{SHOT}&\checkmark&92.4 &92.6 &78.6& 99.0 &76.9 &99.8& 89.9\\
    SHOT-S\dag~\citep{SHOT}&\checkmark&96.0&93.5&81.3&\textbf{99.2}&80.4&\textbf{100.0}&91.7\\
    \shortname{}-T&\checkmark&94.2&91.4&\underline{82.6}&\textbf{99.2}&79.5&\textbf{100.0}&91.2\\
    DATMamba~\citep{DATMamba}&\checkmark&-&-&-&-&-&-&92.4\\
     \shortname-S&\checkmark&97.0&\underline{96.7}&\textbf{83.8}&\underline{99.1}&\textbf{83.4}&\textbf{100.0}&\textbf{93.3}\\
    \bottomrule
    \end{tabular}}
    \label{tab:office_uda}
\end{table}

%% file: domainnet126.tex
\begin{table}[!th]
\centering
\small
\caption{Accuracy(\%) on DomainNet-126 dataset. The ``SF'' column indicates whether a method is source-free ($\checkmark$) or not (\ding{55}). \dag{}: Results from official code reproduction. Best and second-best results are highlighted in \textbf{bold} and \underline{underline}, respectively.}
\setlength{\tabcolsep}{1.pt}

\resizebox{\linewidth}{!}{
\begin{tabular}{l|c|*{8}{c}}
   \toprule
    \multirow{1}{*}{Method} &SF&R$\xrightarrow{}$C& R$\xrightarrow{}$P& P$\xrightarrow{}$C &C$\xrightarrow{}$S &S$\xrightarrow{}$P& R$\xrightarrow{}$S& P$\xrightarrow{}$R & Avg \\
    \midrule
       \multicolumn{10}{c}{\textbf{CNN-based Methods}}\\
   \midrule
    ResNet50~\citep{Resnet}&\checkmark&55.5& 62.7& 53.0& 46.9 &50.1& 46.3& 75.0& 55.6\\
    SHOT ~\citep{SHOT}&\checkmark&67.7&68.4&66.9&60.1&66.1&59.9&80.8&67.1\\
    TENT~\citep{TENT}&\checkmark&58.5 &65.7& 57.9 &48.5& 52.4 &54.0& 67.0 &57.7\\
    AdaCon~\citep{Adacon}&\checkmark&70.2 &69.8& 68.6 &58.0 &65.9& 61.5 &80.5& 67.8\\
    UPA~\citep{UPA}&\checkmark&68.6& 69.5& 67.6& 60.9& 66.8&61.5& 80.9& 68.0\\
    $\text{SF(DA)}^2$~\citep{sfda2}&\checkmark&68.8&70.5&67.8&59.6&67.7&60.2&83.5&68.3\\
    C-SFDA~\citep{C-SFDA}&\checkmark&70.8 &71.1& 68.5 &62.1 &67.4 &62.7& 80.4& 69.0\\
    G-SFDA~\citep{G-SFDA}&\checkmark&74.2& 70.4& 68.8& 64.0& 67.5& 65.7& 76.5& 69.6\\
    RGV~\citep{RGV}&\checkmark&\underline{77.6}& 72.9& 71.9& 68.0& 71.2& 67.3& 83.2& 73.2\\
    \midrule
    \multicolumn{10}{c}{\textbf{ViT-based Methods}}\\
    \midrule
    DeiT-S~\citep{DeiT}&\checkmark&63.1&68.6&61.4&56.5&61.9&53.1&77.2&63.1\\
    DetT-B~\citep{DeiT}&\checkmark&68.1&72.8&69.3&64.4&68.0&59.5&82.8&69.3\\
    CDTrans-S$\dag$~\citep{CDTrans}&\ding{55}&68.3&71.4&70.7&65.1&70.8&60.5&82.0&69.8\\
    CDTrans-B$\dag$~\citep{CDTrans}&\ding{55}&77.0&\underline{75.7}&77.0&72.6&\underline{76.5}&66.1&\textbf{85.9}&75.8\\
    \midrule
    \multicolumn{10}{c}{\textbf{Mamba-based Methods}}\\
    \midrule
    Source-T&\checkmark&67.7&71.0&69.1&66.5&68.3&61.9&81.1&69.4\\
    Source-S&\checkmark& 70.7&73.5&71.2&69.2& 71.0&64.6&82.4&71.8\\
    SHOT-T\dag~\citep{SHOT}&\checkmark&73.2& 73.0 &74.0 &72.4 &72.9& 68.1& 83.3& 73.8\\
    SHOT-S\dag{}~\citep{SHOT}&\checkmark&75.8&\textbf{76.4}&77.0&74.4&76.3&\underline{71.1}&84.6&\underline{76.5} \\
    \shortname-T &\checkmark&76.4 &73.9 &\underline{77.4}& \underline{75.3} &75.3& 71.0 &84.1 &76.2\\
    \shortname-S &\checkmark &\textbf{77.9}&\textbf{76.4}&\textbf{78.8}&\textbf{76.4}&\textbf{77.4}&\textbf{73.5}&\underline{85.2}&\textbf{77.9}\\
    \bottomrule
\end{tabular}}
\label{tab:domainnet126}
\end{table}

%% file: abl_module.tex
\begin{table}[!t]
    \centering
    \small
    \caption{Component ablation study of \shortname{} on Office-Home and VisDA-C. Performance is reported under both the source-only and adaptation settings, with $\text{Baseline}_S$ and $\text{Baseline}_T$ denoting the base models for each scenario, respectively. Best results are in \textbf{bold}.}
    \resizebox{\linewidth}{!}{
    \begin{tabular}{l|cc|cc}
    \toprule
        &\shorta & \shortb & Office-Home & VisDA-C\\
        \midrule
        \multicolumn{5}{c}{\textit{Source-Model Only}}\\
        \midrule    
        $\text{Baseline}_S$ &&-&74.4&67.3\\
        Source-S &\checkmark&-&75.3&65.6\\
        \midrule
        \multicolumn{5}{c}{\textit{After Adaptation}}\\
        \midrule
        $\text{Baseline}_T$&&&77.8&80.9\\
        $w/$ \shortb &&\checkmark&79.0&83.2\\
        $w/$ \shorta & \checkmark&&81.2&85.4\\
        \shortname & \checkmark&\checkmark& \textbf{81.7}&\textbf{89.3}\\
    \bottomrule
    \end{tabular}}
    \label{tab:abl_module}
\end{table}

%% file: abl_attnv2.tex
\begin{table*}[!th]
    \centering
    \caption{Ablations of different channel attention mechanisms on transfer tasks in the Office-Home dataset with only loss $L_{ent}$ and $L_{div}$. Performance of the \textit{source-model-only / adapted target model} is reported. Best and second-best results are highlighted in \textbf{bold} and \underline{underline}, respectively.}
    \small
\setlength{\tabcolsep}{1pt}
    \resizebox{\textwidth}{!}{
    \begin{tabular}{l|*{12}{c}|c}
    \toprule
    Attention & A$\xrightarrow{}$C& A$\xrightarrow{}$P&A$\xrightarrow{}$R& C$\xrightarrow{}$A &C$\xrightarrow{}$P&C$\xrightarrow{}$R &P$\xrightarrow{}$A &P$\xrightarrow{}$C&P$\xrightarrow{}$R& R$\xrightarrow{}$A&R$\xrightarrow{}$C&R$\xrightarrow{}$P&Avg\\
    \midrule
    None&62.2/66.6&77.9/79.1& 83.5/82.9 &\underline{75.6}/77.8&78.6/82.5&\underline{81.5}/81.8&70.7/75.4&56.0/63.3&83.5/85.0&77.4/81.0&59.0/68.0&\underline{86.2}/88.6&74.3/77.7\\
    \midrule
    SA & 62.2\stdvflat/\underline{67.6}\stdvu{}&77.8\stdvddown{}/79.8\stdvu{}& 83.5\stdvflat/85.9\stdvu{}&74.1\stdvddown{}/74.0\stdvddown{}&79.9\stdvu{}/81.1\stdvddown{}&80.2\stdvddown{}/82.4\stdvu{}&69.7\stdvddown{}/71.7\stdvddown{}&55.5\stdvddown{}/\underline{65.0}\stdvu{}&83.2\stdvddown{}/85.5\stdvu{}&77.0\stdvddown{}/80.8\stdvddown{}&\underline{60.5}\stdvu{}/68.2\stdvu{}&85.7\stdvddown{}/88.5\stdvddown{}&74.1\stdvddown{0.2}/77.5\stdvddown{0.2}\\
       CM&\textbf{63.0}\stdvu{}/66.3\stdvddown{}&\underline{78.2}\stdvu{}/79.0\stdvddown{}&84.0\stdvu{}/\underline{86.4}\stdvu{}&\textbf{76.2}\stdvu{}/74.1\stdvddown{}&79.1\stdvu{}/\underline{83.4}\stdvu{}&81.0\stdvddown{}/\underline{84.5}\stdvu&69.8\stdvddown{}/74.1\stdvddown{}&55.6\stdvddown{}/64.9\stdvu{}&83.9\stdvu{}/\underline{86.2}\stdvu{}&\underline{77.8}\stdvu{}/80.8\stdvddown{}&59.7\stdvu{}/\underline{69.3}\stdvu{}&85.9\stdvddown{}/\underline{89.2}\stdvu{}&74.5\stdvu{0.2}/78.2\stdvu{0.5}\\
       ECA &62.8\stdvu{}/66.0\stdvddown{}&\underline{78.2}\stdvu{}/79.3\stdvu{}&83.7\stdvu{}/82.0\stdvddown{}&74.6\stdvddown{}/75.6\stdvddown{}&79.7\stdvu{}/80.9\stdvddown{}&81.0\stdvddown{}/79.2\stdvddown{}&70.0\stdvddown{}/74.2\stdvddown{}&56.8\stdvu{}/62.3\stdvddown{}&83.4\stdvddown{}/84.0\stdvddown{}&76.4\stdvddown{}/80.6\stdvddown{}&60.4\stdvu{}/68.2\stdvu{}&85.5\stdvddown{}/87.1\stdvddown{}&74.4\stdvu{0.1}/76.6\stdvddown{1.1}\\
       CBAM& 62.5\stdvu{}/65.1\stdvddown{}&78.0\stdvu{}/\underline{81.0}\stdvu{}&\underline{84.1}\stdvu{}/84.4\stdvu{}&74.5\stdvddown{}/76.4\stdvddown{}&\underline{80.1}\stdvu{}/82.1\stdvddown{}&\textbf{81.8}\stdvu{}/81.6\stdvddown{}&70.7\stdvflat/74.7\stdvddown{}&\textbf{57.2}\stdvu{}/63.3\stdvflat&\underline{84.0}\stdvu{}/85.0\stdvflat&77.3\stdvddown{}/\underline{81.1}\stdvu{}&59.9\stdvu{}/68.6\stdvu{}&85.4\stdvddown{}/88.6\stdvflat&74.6\stdvu{0.3}/77.7\stdvflat{0.0}\\
      \midrule
      Ch-Group &62.7\stdvu{}/66.2\stdvddown{}&\textbf{79.5}\stdvu{}/80.8\stdvu{}&\textbf{84.7}\stdvu{}/\textbf{86.9}\stdvu{}&75.3\stdvddown{}/\underline{79.9}\stdvu{}&\textbf{80.2}\stdvu{}/\textbf{85.0}\stdvu{}&81.2\stdvddown{}/\textbf{84.8}\stdvu{}&\underline{71.8}\stdvu{}/\underline{77.1}\stdvu{}&\underline{56.9}\stdvu{}/\textbf{66.4}\stdvu{}&\textbf{84.4}\stdvu{}/\textbf{86.6}\stdvu{}&\textbf{78.4}\stdvu{}/\textbf{81.3}\stdvu{}&\textbf{61.9}\stdvu{}/\textbf{70.1}\stdvu{}&\textbf{86.6}\stdvu{}/\textbf{89.7}\stdvu{}&\textbf{75.5}\stdvu{1.2}/\underline{79.2}\stdvu{1.5}\\
       \shorta &\underline{62.9}\stdvu{}/\textbf{68.7}\stdvu{}&\textbf{79.5}\stdvu{}/\textbf{83.2}\stdvu{}&\textbf{84.7}\stdvu{}/\textbf{86.9}\stdvu{}&74.4\stdvddown{}/\textbf{80.4}\stdvu{}&\textbf{80.2}\stdvu{}/\textbf{85.0}\stdvu{}&81.2\stdvddown{}/\textbf{84.8}\stdvu{}&\textbf{71.9}\stdvu{}/\textbf{78.7}\stdvu{}&\underline{56.9}\stdvu{}/\textbf{66.4}\stdvu{}&\textbf{84.4}\stdvu{}/\textbf{86.6}\stdvu{}&\textbf{78.4}\stdvu{}/\textbf{81.3}\stdvu{}&\textbf{61.9}\stdvu{}/\textbf{70.1}\stdvu{}&\textbf{86.6}\stdvu{}/\textbf{89.7}\stdvu{}&\underline{75.3}\stdvu{1.0}/\textbf{80.2}\stdvu{2.5}\\
    \bottomrule
    \end{tabular}}
    \label{tab:abl_attn}
\end{table*}

%% file: abl_backbone.tex
\begin{table*}[!t]
    \centering
    \caption{\srevise{Ablation study of our \shorta{} module on the source-model only and after adaptation (with only loss $L_{ent}$ and $L_{div}$) performance under different backbones on the Office-Home dataset. Best and second-best results on each stage are highlighted in \textbf{bold} and \underline{underline}, respectively.}}
    \srevise{
    \resizebox{.98\textwidth}{!}{
    \begin{tabular}{l|c|c|*{13}{c}}
    \toprule
        Backbones&Params&\textbf{\shorta}& A$\xrightarrow{}$C& A$\xrightarrow{}$P& A$\xrightarrow{}$R& C$\xrightarrow{}$A& C$\xrightarrow{}$P& C$\xrightarrow{}$R& P$\xrightarrow{}$A& P$\xrightarrow{}$C& P$\xrightarrow{}$R& R$\xrightarrow{}$A& R$\xrightarrow{}$C& R$\xrightarrow{}$P& Avg\\
        \midrule
        \multicolumn{16}{c}{\textit{Source-Model Only}}\\
        \midrule    
        \multirow{2}{*}{Agent-DeiT-B~\citep{agent_attention}}&\multirow{2}{*}{87M}&\ding{55}&56.4&71.3&79.4&69.5&72.5&	76.5&62.5&52.4&79.7&74.2&54.8&84.1&69.4\\
     &&\checkmark&61.4\stdvu{}&77.2\stdvu{}&	81.7\stdvu{}&67.7\stdvddown{}&75.9\stdvu{}&77.1\stdvu{}&66.3\stdvu{}&55.0\stdvu{}&81.4\stdvu{}&73.8\stdvddown{}&	57.1\stdvu{}&83.7\stdvddown{}&71.5\stdvu{2.1}\\   
        \midrule
        \multirow{2}{*}{DeiT-B~\citep{DeiT}}&\multirow{2}{*}{86M}&\ding{55}&58.2&72.2&79.3&	67.7&71.8&75.6&61.4&50.4&79.1&70.7&	54.6&81.6&68.6\\
        &&\checkmark&57.3\stdvddown{}&74.6\stdvu{}&81.0\stdvu{}&68.7\stdvu{}&74.5\stdvu{}&75.9\stdvu{}&65.0\stdvu{}&	51.8\stdvu{}&81.4\stdvu{}&72.4\stdvu{}&56.1\stdvu{}&82.0\stdvu{}&70.1\stdvu{1.5}\\
        \midrule
        \multirow{2}{*}{ConvNeXt-B~\citep{convnext}}&\multirow{2}{*}{89M}&\ding{55}&61.5	&76.2&82.2&71.6&76.3&79.0&69.8&	55.3&82.7&77.5&	\underline{62.2}&85.0&73.3\\
        && \checkmark&\textbf{64.6}\stdvu{}&\underline{79.0}\stdvu{}&\underline{83.5}\stdvu{}&\underline{74.9}\stdvu{}&\underline{78.7}\stdvu{}&81.1\stdvu{}&\underline{71.8}\stdvu{}&\textbf{57.3}\stdvu{}&	\textbf{84.5}\stdvu{}&77.9\stdvu{}&\textbf{62.7}\stdvu{}&85.9\stdvu{}&\underline{75.2}\stdvu{1.9}\\
        \midrule
        \multirow{2}{*}{Mamba-S}&\multirow{2}{*}{50M}&\ding{55}&62.2&77.7&83.4&\textbf{75.2}&\textbf{80.2}&	\textbf{82.0}&71.2&\underline{57.1}&83.6&\underline{78.3}&61.4&\underline{86.5}&74.9\\
        &&\checkmark&\underline{62.9}\stdvu{}&\textbf{79.5}\stdvu{}&\textbf{84.7}\stdvu{}&74.4\stdvddown{}&\textbf{80.2}\stdvu{}&\underline{81.2}\stdvddown{}&\textbf{71.9}\stdvu{}& 56.9\stdvddown{}&\underline{84.4}\stdvu{}&\textbf{78.4}\stdvu{}&61.9\stdvu{}& \textbf{86.6}\stdvu{}&\textbf{75.3}\stdvu{0.4}\\
        \midrule
        \multicolumn{16}{c}{\textit{After Adaptation}}\\
        \midrule
        \multirow{2}{*}{Agent-DeiT-B~\citep{agent_attention}}&\multirow{2}{*}{87M}&\ding{55}&63.1&72.9&80.6&70.5&73.6&76.2&66.4&	58.5&81.0&77.4&64.9&85.9&72.6 \\
        &&\checkmark&64.6\stdvu&81.3\stdvu&84.7\stdvu&74.7\stdvu&80.9\stdvu&80.5\stdvu&74.1\stdvu&62.2\stdvu&	\underline{85.2}\stdvu&78.5\stdvu&66.8\stdvu&88.6\stdvu&76.8\stdvu{4.2}\\
        \midrule
        \multirow{2}{*}{DeiT-B~\citep{DeiT}}&\multirow{2}{*}{86M}&\ding{55}&61.0&76.3&81.6&69.8&73.6&78.6&65.8&58.8&81.4&76.6&61.8&84.8&72.5 \\
        &&\checkmark&64.2\stdvu&78.7\stdvu&83.5\stdvu&	72.8\stdvu&80.6\stdvu&79.8\stdvu&72.8\stdvu&60.6\stdvu&83.5\stdvu&77.6\stdvu&65.4\stdvu&87.5\stdvu&75.6\stdvu{3.1}\\
        \midrule
        \multirow{2}{*}{ConvNeXt-B~\citep{convnext}}&\multirow{2}{*}{89M}&\ding{55}&62.9&76.9&83.3&75.4&78.1&78.5&72.3&60.3&82.3&76.3&	62.6&82.8&74.3\\
        &&\checkmark&\underline{68.1}\stdvu&\underline{81.5}\stdvu&\textbf{87.5}\stdvu&\underline{79.9}\stdvu&\underline{82.7}\stdvu&\underline{83.2}\stdvu&\underline{77.5}\stdvu&\underline{65.8}\stdvu&	\textbf{86.6}\stdvu&80.3\stdvu&\textbf{70.2}\stdvu&\underline{88.7}\stdvu&\underline{79.3}\stdvu{5.0}\\
        \midrule
        \multirow{2}{*}{VMamba-S}&\multirow{2}{*}{50M}&\ding{55}&64.0&78.2&82.9&78.1&82.5&81.8&75.5&63.3&85.0&\underline{81.0}&68.0&88.6&77.4\\
        &&\checkmark&\textbf{68.7}\stdvu&\textbf{83.2}\stdvu&\underline{86.9}\stdvu&\textbf{80.4}\stdvu&\textbf{85.0}\stdvu&\textbf{84.8}\stdvu&\textbf{78.7}\stdvu&	\textbf{66.4}\stdvu&\textbf{86.6}\stdvu&\textbf{81.3}\stdvu&\underline{70.1}\stdvu&\textbf{89.7}\stdvu&\textbf{80.2}\stdvu{2.8}\\
    \bottomrule
    \end{tabular}}}
    \label{tab:abl_backbone}
\end{table*}

%% file: abl_scs.tex
\begin{table}[!t]
    \centering
    \caption{Ablation on the background (Bg) identification strategy in \shortb: our selective shuffling vs. a random shuffling baseline.}
    \resizebox{\linewidth}{!}{
    \begin{tabular}{l|c|*{4}{c}|c}
    \toprule
         \multirow{2}{*}{}&\multirow{2}{*}{Bg}&\multicolumn{4}{c|}{Office-Home}&VisDA-C\\
         \cmidrule{3-7}
&&A$\xrightarrow{}$C&A$\xrightarrow{}$P&C$\xrightarrow{}$A&P$\xrightarrow{}$A &S$\xrightarrow{}$R \\
         \midrule
         None&-&70.3&83.8&81.8&78.7&85.4\\
         Random&\ding{55}&70.8&84.4&\textbf{82.0}&78.7&88.5\\
\shortb&\checkmark&\textbf{71.2}&\textbf{84.5}&81.7&\textbf{79.0}&\textbf{89.3}\\
    \bottomrule
    \end{tabular}}
    \label{tab:abl_scs}
\end{table}

%% file: flops.tex
\begin{table}[!t]
    \centering
    \caption{Computational efficiency comparison between SOTAs and our \shortname{} on the Office-Home (OH) and VisDA-C (V-C) datasets. Evaluations were conducted using 224×224 resolution images, with throughput measured at batch size 16 and FLOPs calculated at batch size 1 on an NVIDIA A100 GPU. Key metrics: P = Parameters (M), F = FLOPs (G), Th = Throughput (images/s).}
\setlength{\tabcolsep}{1pt}
    \resizebox{\linewidth}{!}{

    \begin{tabular}{ll|cccccc}
    \toprule
Methods&Model&P(M)&F(G)$\downarrow$&Th$\uparrow$&OH(\%)$\uparrow$ &V-C(\%)$\uparrow$\\
    \midrule
         I-SFDA~\citep{I-SFDA}&ResNet101&43.0&7.8&1094.7&-&88.4\\
         Ours&VMamba-T&39.0&5.4&700.9 &\textbf{79.2}&\textbf{88.5}\\
         \midrule
         CDTrans~\citep{CDTrans}&DeiT-B&86.6&17.6 &399.5&80.5&88.4\\
         DATMamba~\citep{DATMamba}&Mamba&97.6&15.0& -&80.9&88.9\\
         Ours&VMamba-S&58.9&9.2&502.8 &\textbf{81.7}&\textbf{89.3}\\
    \bottomrule
    \end{tabular}}
    \label{tab:flops}
\end{table}

%% file: suppl_abl_module.tex
\begin{table}[!t]
    \centering
    \small
    \caption{Ablation study on Office-Home and VisDA-C datasets, reporting target domain adaptation performance.}
    \begin{tabular}{l|cc|cc}
    \toprule
         &Filter & \shortb & Office-Home & VisDA-C\\
        \midrule
        Baseline\ddag &&&80.3&79.9\\
        $w/$ SCS & &\checkmark&80.9&89.3\\
        $w/$ Filter & \checkmark&&81.2&85.4\\
        \shortname & \checkmark&\checkmark& 81.7&89.3\\
    \bottomrule
    \end{tabular}
    \label{tab:suppl_abl_module}
\end{table}